\documentclass[pdflatex,sn-basic]{sn-jnl}

\usepackage{graphicx}
\usepackage{amsmath,amssymb,amsfonts}
\usepackage{multirow}
\usepackage{placeins}
\usepackage{chngcntr}
\usepackage{textcomp}
\usepackage{xcolor}
\usepackage{url}
\let\orcidlogo\relax
\usepackage{orcidlink}

\begin{document}

\title[Beyond Point Prediction]{Beyond Point Prediction: Artificial Representative Trees with Uncertainty}

\author*[1]{\fnm{Lea L.} \sur{Mairhöfer} \orcidlink{0000-0002-2126-6831}}\email{lea.mairhoefer@uni-luebeck.de}
\author[1]{\fnm{Silke} \sur{Szymczak}\orcidlink{0000-0002-8897-9035}}
\author[1,2,3]{\fnm{Björn-Hergen} \sur{Laabs}\orcidlink{0000-0002-9265-5738}}
\author[4]{\fnm{Tuwe} \sur{Löfström-Cavallin}\orcidlink{0000-0003-0274-9026}}

\affil*[1]{\orgdiv{Institute of Medical Biometry and Statistics}, \orgname{University of Luebeck}, \orgaddress{\street{Ratzeburger Allee 160 V24}, \postcode{23562}, \city{Lübeck}, \country{Germany}}}
\affil[2]{\orgdiv{Department of Medical Statistics}, \orgname{University Medical Center Göttingen}, \orgaddress{\city{Göttingen}, \country{Germany}}}
\affil[3]{\orgname{CAIMed, Lower Saxony Center for Artificial Intelligence and Causal Methods in Medicine}, \orgaddress{\city{Göttingen}, \country{Germany}}}
\affil[4]{\orgdiv{Department of Computing}, \orgname{Jönköping University}, \orgaddress{\city{Jönköping}, \country{Sweden}}}

\abstract{
    Random forests (RFs) predict well but are opaque, whereas single decision trees are interpretable but unstable. Artificial representative trees (ARTs) were developed as interpretable surrogate models for RFs, but their use as standalone prediction models with uncertainty quantification has not been systematically investigated. We combine ARTs with leaf-wise Mondrian conformal predictive systems (CPS), enabling a single tree to provide continuous predictions, prediction intervals, and probabilities of exceeding arbitrary thresholds.

	We compared ARTs with CPS against decision trees with CPS and separate regression and probability trees across five simulation scenarios, 21 benchmark datasets, and a cross-sectional NHANES example data set. Repeated cross-validation assessed predictive performance, interpretability, and stability.

	ARTs with CPS yield compact, structurally stable trees with substantially more reproducible split-variable selection than decision trees across benchmark datasets and NHANES. Decision trees showed slightly better predictive performance and narrower prediction intervals, while coverage was broadly comparable. CPS-based trees generally achieved lower and less variable Brier scores than multi-model approaches.

	Combining ARTs with CPS therefore provides a single, interpretable, and stable model for continuous predictions and calibrated probabilities, balancing predictive performance with reproducibility and transparency in settings where stability and interpretability are essential.
}



\maketitle

\section{Introduction}\label{sec:introduction}
Random forests (RFs) often achieve strong predictive performance by aggregating many decorrelated decision trees, but the resulting ensemble does not provide a single set of decision rules that can be inspected directly \citep{bostrom_explaining_2018, molnar_interpretable_2022}. 
A single decision tree would provide this transparency, yet its structure can change markedly by small changes to the training data. Such structural instability is particularly problematic when a tree is used for interpretation. Conclusions about relevant variables and decision paths may then depend on the particular sample or resampling fold rather than on a reproducible pattern in the data. 

An ART is constructed from the most important splits observed in an RF and results in a compact tree that is representative of the ensemble according to a chosen tree-distance measure \citep{laabs_construction_2024}. ARTs were originally proposed as surrogate models for explaining RFs. 
Here, we study a different use. The ART itself serves as a standalone prediction model. The RF is only used during the construction phase to aggregate information about potential splits. Predictions are then obtained from the ART directly.

Despite the stability of the model in many high-stakes decisions (e.g., medical application) a measure of prediction uncertainty is required. Like decision trees, ARTs only provide point predictions for an individual outcome value. 
Conformal prediction provides a model-agnostic framework for constructing prediction sets under exchangeability, with finite-sample marginal coverage determined by the calibration procedure \citep{vovk_algorithmic_2005}. 
Conformal predictive systems (CPS) extend this idea from prediction sets to predictive distributions, from which point predictions, prediction intervals, and probabilities of exceeding any prespecified threshold can be derived \citep{vovk_nonparametric_2019}. 
Previous work has combined conformal prediction with decision trees \citep{johansson_conformal_2013,johansson_interpretable_2018}. For classification, inductive conformal prediction with unpruned trees and smoothed probability estimates yielded efficient prediction sets while preserving a single interpretable tree \citep{johansson_conformal_2013}. For regression, leaf-wise normalization retained rule-based interpretability, while Mondrian calibration additionally provided validity for each leaf-specific rule, at the cost of wider prediction intervals, particularly at very high confidence levels \citep{johansson_interpretable_2018}.
Mondrian conformal methods partition calibration observations into predefined categories and calibrate within each category \citep{bostrom_mondrian_2020,bostrom_mondrian_2021}. 
When tree leaves define the categories, all observations following the same decision path receive uncertainty estimates based on the corresponding leaf. 
This preserves the rule-based interpretation of the tree, although category-specific validity requires exchangeability within categories and sufficiently many calibration observations per leaf.

We therefore combine an ART with a leaf-wise Mondrian CPS. 
The resulting single model predicts a continuous outcome, provides corresponding prediction intervals, and estimates probabilities that the outcome exceeds clinically relevant or user-defined thresholds.
With this approach, it is possible to characterize and diagnose patients in a unified, uncertainty-aware framework without the need to estimate multiple models on medical data. 
Our primary research question is whether this approach improves structural stability, particularly the reproducibility of selected splitting variables across folds and repetitions, while retaining useful predictive performance and interpretable decision rules.
We compare ARTs with CPS against conventional decision trees with CPS and against ARTs and decision trees fitted separately for continuous and threshold-based outcomes. 

The evaluation comprises five simulation scenarios and 21 benchmark regression datasets selected from earlier studies, following recommendations for neutral method-comparison studies \citep{boulesteix_plea_2013,friedrich_role_2024}. 
We therefore assess performance on both the simulated datasets from the original ART publication \cite{laabs_construction_2024} and the benchmark datasets used in the publication introducing decision trees with conformal prediction \citep{johansson_conformal_2013}.
We assess predictive accuracy, interval coverage and width, tree size, and several measures of structural stability. 
Because ART construction involves choices such as the distance metric, the minimum leaf size (\emph{min.bucket}), and the restriction of candidate split points (\emph{probs\_quantiles}), we additionally examine how these hyperparameters affect prediction, stability, and interpretability.

Finally, we illustrate the fitted model using cross-sectional data from the National Health and Nutrition Examination Survey (NHANES). The example predicts glycohemoglobin and the probabilities of exceeding the commonly used values 5.7 and 6.5 for prediabetes and diabetes \citep{casacchia_development_2024}. 
It is intended solely to demonstrate how continuous predictions, uncertainty intervals, threshold probabilities, and decision paths can be presented together. 
Owing to the cross-sectional study design, it is not proposed or evaluated as a clinical prediction model, and no claims about prospective clinical utility are made. 
Together, the simulation, benchmark, and illustrative analyses are designed to establish whether ARTs with CPS offer a reproducible and transparent alternative to conventional single-tree prediction.

\section{Background}\label{sec:background}
\subsection{Decision tree}
A decision tree recursively partitions the predictor space into disjoint regions and assigns a prediction to each leaf \citep{breiman_classification_1984}. Starting at the root node, the algorithm greedily selects the variable and split point that optimize a predefined splitting criterion. For regression, this criterion commonly corresponds to a reduction in the within-node variance. 
The procedure is repeated within the resulting child nodes until a stopping rule is reached, for example a minimum number of observations in a node. 
The path from the root to a leaf forms an explicit decision rule, making both the model structure and individual predictions directly inspectable.

This interpretability comes at the cost of instability, as decision trees tend to overfit \citep{breiman_random_2001}. Consequently, RFs are commonly preferred, as they are more stable and, therefore, have better prediction performance on test data.

\subsection{Random forest (RF)}
An RF consists of an ensemble of decision trees \citep{breiman_random_2001}. Each tree is grown on a bootstrap sample of the training data to increase stability by decorrelating the individual trees. In addition, at each split only a random subset of candidate variables is considered for splitting. 
In realistic scenarios, RFs typically consist of many deep trees, making the decision paths that lead to an individual prediction difficult to understand for humans. As a result, their predictions are often hard to interpret \citep{bostrom_explaining_2018, molnar_interpretable_2022}, so that a plausibility check of the model is difficult. 

Although feature importance measures, including inferential procedures for high-dimensional data, can be computed to assess the relevance of variables \citep{janitza_computationally_2018}, the tree structure is not captured.
The same applies to the use of post hoc methods such as partial dependence plots (PDP) or shapley additive explanations (SHAP) \citep{friedman_greedy_2001, lundberg_unified_2017}.
Although these approaches can highlight the marginal effects of individual variables (PDP) or assign additive importance scores to input variables based on cooperative game theory (SHAP), they neither display the decision paths within the ensemble nor can be used as prediction models on their own.

\subsection{Artificial representative tree (ART)}
A popular method for interpreting complex models is the use of interpretable surrogate models \citep{banerjee_identifying_2012,craven_extracting_1995,guidotti_survey_2018,johansson_accurate_2014,molnar_interpretable_2022}. 
This approach is  well established and has been applied in various contexts, including rule extraction to handle the tradeoff between prediction performance and comprehensibility \citep{johansson_why_2006}.
A simpler and more interpretable model is created to approximate the structure of the complex model as closely as possible. Once the surrogate model is generated, it can be analyzed and interpreted to gain insights into the behavior and decisions of the original, complex model.
ARTs are a special type of surrogate models that aim to maintain the stability of the RF while producing a single, small, and easily understandable tree \citep{laabs_construction_2024}.

To construct an ART, a new tree is generated using only splits that occur in the RF. 
Using these split points, all possible stumps are created. The pairwise distance between each stump and all trees in the RF is then measured, for which various distance metrics can be used (hyperparameter: \emph{metric}). For example, one can maximize similarity to the RF in terms of predictions (prediction), the split variables used (splitting variables, SV), or an extended version of the split variables that is weighted by the position of their appearance in the tree (weighted splitting variables, WSV).
The stump with the lowest mean distance to all trees in the RF is used to continue growing the ART. Next, all possible trees with one additional split are constructed. The pairwise distance between each candidate tree and all trees in the RF is computed and the tree with the smallest mean distance is chosen again. 
This process of adding splits continues iteratively until the mean distance no longer decreases. 

To reduce runtime, it is possible to restrict the RF splits for continuous variables to selected quantiles (hyperparameter: \emph{probs\_quantiles}).
To control tree growth, users can specify a minimum number of observations per node (hyperparameter: \emph{min.bucket}). Splits are only added to a node as long as the number of training observations in each resulting child node does not fall below this threshold. 

In the original surrogate setting, an ART is evaluated by how well it represents the RF. In contrast, the present study uses the ART as the final prediction model. The RF is only used during training to generate an ART. Accordingly, point predictions, prediction errors, prediction intervals, and threshold probabilities are evaluated with respect to the ART and the observed outcomes, not with respect to RF predictions. 

As a standalone model, the resulting ART is small, stable and thus easy to interpret.  Like an ordinary regression tree, however, an ART initially provides only point predictions.

\subsection{Conformal predictive systems (CPS)}
Predictive uncertainty can be quantified in several ways, including parametric or Bayesian predictive distributions, ensemble and resampling methods, and direct estimation of conditional quantiles \citep{hullermeier_aleatoric_2021}. 
Tree-based approaches have also been developed to construct prediction intervals for conventional and streaming data settings \citep{krzanowski_recursive_2007,zhao_interval_2021}.
However, obtaining both continuous outcome predictions and probabilities of exceeding a predefined threshold would require fitting separate models.
Parametric and Bayesian approaches provide predictive distributions but rely on distributional assumptions, ensemble methods estimate uncertainty from variation across multiple models but do not preserve the interpretation of a single tree \citep{lakshminarayanan_simple_2016}, and quantile-based approaches generally do not provide finite-sample coverage guarantees \citep{meinshausen_quantile_2006}. We therefore use CPS, which provide a distribution-free and model-agnostic framework for predictive uncertainty quantification with finite-sample marginal validity under exchangeability \citep{vovk_nonparametric_2019}. This is particularly suitable for our objective, as a single fitted and interpretable ART can provide point predictions, prediction intervals at different coverage levels, and probabilities of exceeding arbitrary clinically relevant thresholds.

For uncertainty quantification in regression tasks, CPS extends point predictions to one- or two-sided prediction intervals that correspond to a chosen coverage $1-\alpha$ for the significance level $\alpha$. Unlike confidence intervals, which quantify uncertainty around parameter estimates, prediction intervals describe the range within an observation is expected to fall, with a specified confidence based on the selected significance level. 

Conformal prediction can be implemented using either an inductive or a transductive approach \citep{johansson_conformal_2013}. For inductive settings, the model is trained and calibrated once, while for transductive settings, the system is retrained for each test observation to improve validity but at greater computational cost.
Here, only the inductive setting is used, as it is more efficient and allows the uncertainty to be computed once per ART or decision tree to aid interpretation.

To develop a model with uncertainty quantification using CPS, the available data is first divided into training, calibration, and test datasets. Subsequently, the following steps are performed:
\begin{enumerate}
    \item Train a machine learning model using the training data.
    \item Use the model to compute predictions $\hat{y}_{(i)}^c$ for all observations $i$ in the calibration dataset, with $i$ ranging from 1 to the number of observations $q$ in the calibration dataset.
    \item  Residuals are calculated using a nonconformity function to measure how much the true outcome $y_{(i)}^c$ of the observation $i$ deviates from the predictions of the model. For regression, usually the residuals $\epsilon_{(i)} = y_{(i)}^c-\hat{y}_{(i)}^c$ are used. The residuals are then sorted in ascending  order.
    \item Add the residuals to the point prediction of a test observation $t$, resulting in $\mathbb{C}_{(i)}=\hat{y}_{t} + \epsilon_{(i)}$. The first and last value of $\mathbb{C}_{(i)}$ are defined as $\mathbb{C}_{(0)}=-\infty$ and $\mathbb{C}_{(q+1)}=\infty$. 
    \item Choose the significance level $\alpha\in (0,1)$.
    \item Calculate the prediction interval for the predictions of the test data. For a given significance level $\alpha$, a two-sided prediction interval is chosen as $[\mathbb{C}_{\lfloor(\alpha/2)(q+1)\rfloor}, \mathbb{C}_{\lceil(1-\alpha/2)(q+1)\rceil}]$ and, analogously, the one-sided intervals are defined as $[-\infty, \mathbb{C}_{\lceil(1-\alpha)(q+1)\rceil}]$ for an upper bounded interval or $[\mathbb{C}_{\lfloor(\alpha)(q+1)\rfloor}, \infty]$ for a lower bounded interval. 
\end{enumerate}
The minimum number of observations required in the calibration set can be derived as $\alpha^{-1}-1$. This result originates from conformal prediction theory and can be transferred to CPS \citep{johansson_interpretable_2018}.

In addition to a prediction interval, an empirical cumulative distribution can be constructed, called the conformal predictive distribution. This distribution can be used to estimate the probability that the true outcome is below a given threshold $\theta$. The conformal predictive distribution is defined as
\begin{align*}
        \mathbb{Q}(\theta)= \begin{cases}
           \frac{i + \tau}{q+1} \text{, if } \theta \in (\mathbb{C}_{(i)}, \mathbb{C}_{(i+1)})& \text{for } i \in {0,\dots,q}\\
            \frac{i'-1+(i''-i'+2)\tau}{q+1}\text{, if } \theta = \mathbb{C}_{(i)}& \text{for } i \in {1,\dots,q}
      \end{cases}
\end{align*} where $\tau$ is drawn from the uniform distribution $U(0, 1)$ so that the p-values of the CPS are uniformly distributed.
$i''$ is the highest index, where $\theta = \mathbb{C}(i'')$, whereas $i'$ is the lowest index, where $\theta = \mathbb{C}(i')$.
In CPS, the p-values quantify how well a new prediction conforms to the distribution of the calibration data.
They are not classical hypothesis testing p-values, but are instead defined by their relative rank among calibration conformity scores. Under exchangeability, they are uniformly distributed between 0 and 1 
Therefore, $P(y \leq \theta)=\mathbb{Q}(\theta)$ indicates the proportion of calibration residuals that are less than or equal to the threshold $\theta$, with a small randomization to account for ties and to maintain uniformly distributed p-values.

As an extension, Mondrian CPS can also be used \citep{bostrom_mondrian_2020, bostrom_mondrian_2021}. The procedure is analogous to CPS, with the difference that the data is divided into disjunct groups and then the CPS is built within each group separately. In the case of decision trees, the groups can be assigned via the leaves \citep{johansson_conformal_2013}. For each leaf, only the observations of the calibration dataset that end in this node are then used to create a prediction interval. This can result in the prediction intervals of the individual leaves having different widths. 
The idea here is that observations that end up in a group, or here in a leaf, are similar to each other and it is therefore better to use only these observations for the respective calibration. The disadvantage, however, is that fewer observations are then available for each calibration calculation. Consequently, the minimum number of calibration observations, must be satisfied within each leaf to ensure valid coverage locally.

\section{Methods}\label{sec:methods}
\subsection{ARTs with uncertainty quantification}
We combine an ART with an inductive leaf-wise Mondrian CPS to obtain an interpretable and stable prediction model that jointly predicts a continuous outcome and the probability of exceeding a pre-defined threshold relevant for diagnosis of a particular disease. Thus, a separate binary model is not required for each threshold.

The procedure uses disjoint training, calibration, and test sets. An RF is fitted using only the training set, and an ART is constructed from the splits of this RF as described above. The RF is used only to construct the ART, all predictions subsequently evaluated in this study are generated by the ART. Once constructed, the ART and its decision rules remain fixed.
Then, we use the calibration set to apply Mondrian CPS to the ART on a leaf-by-leaf basis to compute prediction intervals and cumulative distribution functions for each leaf. 

For a new observation, the ART first assigns the observation to a leaf according to its decision rules. The corresponding leaf determines the point prediction. Its leaf-specific Mondrian calibration is used to derive prediction intervals and probabilities of exceeding one or more user-defined thresholds.

\subsection{R implementation}
The ART implementation has been implemented in our R package \emph{timbR}\footnote{\url{https://github.com/imbs-hl/timbR}}. In version 3.3, we integrated CPS for regression for uncertainty quantification when fitting an ART, a decision tree or a RF using \emph{ranger} \cite{wright_ranger_2017}. For single decision trees or ARTs, Mondrian CPS is additionally available, providing individual prediction intervals for each leaf. Furthermore, the cumulative predictive distribution allows the calculation of probabilities of exceeding a user-defined threshold. Analogous to uncertainty quantification, a Mondrian variant is also available.
To ensure that a sufficient number of observations is available in each leaf of an ART for uncertainty quantification, we added the hyperparameter \emph{min.bucket}, which specifies the minimum number of training observations required in each leaf during ART construction and can be adjusted manually.

CPS has also been integrated into the plotting function, enabling graphical extraction of a decision tree or ART from R, including point predictions, prediction intervals, their empirical coverage, and interval width within each leaf. In addition, the cumulative predictive distribution can be visualized below each leaf together with the probability of exceeding a user-defined threshold.

\begin{figure*}[t]
	\includegraphics[width=\textwidth]{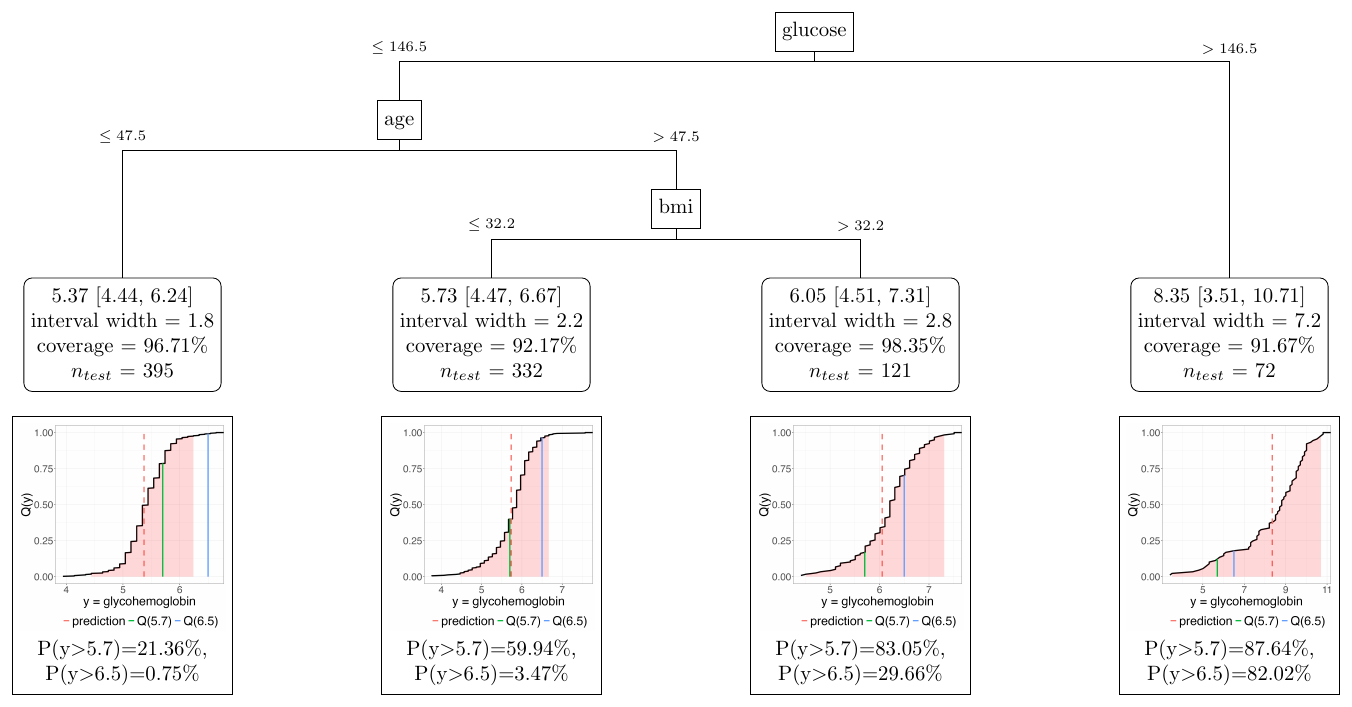}
	\caption{An ART with CPS trained on the NHANES dataset is displayed, including the cumulative distribution function. For each leaf, the point prediction of glycohemoglobin (red dashed line) and the corresponding 95\% prediction interval (red shaded area) are shown. The coverage and interval width are reported, along with the number of test observations on which the coverage is based. The predicted probabilities for prediabetes (green) and diabetes (red) are also provided. \label{fig:art_sv}} 
\end{figure*}

An example of an ART with Mondrian CPS used for uncertainty quantification is shown in Figure \ref{fig:art_sv}.
The ART was fitted to predict glycohemoglobin as outcome $y$, as well as the probabilities $P(y>5.7)$ as the diagnostic threshold for prediabetes and $P(y>6.5)$ as the threshold for diabetes. One half of the data was used for training, one quarter for calibration, and one quarter for testing. Details on the dataset are provided in Section \ref{sec:application_nhanes}.
The ART was trained using the SV distance measure and a \emph{min.bucket} value of 150. It is important to note that the \emph{min.bucket} of 150 refers to the training data when building the ART; therefore, fewer than 150 calibration or test observations may be present in individual leaves. No restrictions were imposed on the selection of split points from the RF distribution via the hyperparameter \emph{probs\_quantiles}, but the number of splits was limited to three.
In each leaf, the point prediction of the continuous outcome glycohemoglobin is reported together with its 95\% prediction interval and the corresponding interval width (see Figure \ref{fig:art_sv}). In addition, the empirical coverage based on the test observations is provided, along with the number of test instances reaching the respective leaf.
The figure below each leaf displays the cumulative probability distribution. The 95\% prediction interval is highlighted, and the point prediction for glycohemoglobin is marked. Furthermore, the probabilities of exceeding the clinical thresholds for prediabetes (5.7) and diabetes (6.5) are reported and visualized.

As illustrated in Figure \ref{fig:art_sv}, interval widths and empirical coverage differ between leaves, enabling leaf-specific uncertainty quantification. Moreover, the predictive distributions vary across leaves, allowing for individualized probability estimates. 

\section{Simulation study}\label{sec:simulation_study} 
\subsection{Aim}
The aim of the simulation study is to investigate whether a single ART with Mondrian CPS can provide interpretable and stable predictions for both a continuous outcome and the probability of exceeding a given threshold.

\subsection{Data generation mechanism}
The simulation is based on the simulated datasets generated in the original publication where ARTs were first introduced \cite{laabs_construction_2024}.
Five different scenarios are considered, each containing $p=100$ variables and a continuous outcome. For each scenario, training, test, and calibration datasets with 1,000 observations each are simulated. In each scenario, one aspect is systematically varied in order to investigate its influence. All simulated outcomes $y$ are standardized to the interval $[0, 1]$.

\paragraph*{Scenario 1 (few large effects)}
All variables are independently and identically distributed (iid) according to a Bernoulli distribution, such that $x_j \overset{\text{iid}}{\sim} \mathcal{B}(0.5)$ for $j = 1, \dots, p$.
The outcome is simulated using five effect variables $x_1, ..., x_5$, each with an effect size of $\beta = 2$, and standard normally distributed noise $\epsilon \sim \mathcal{N}(0,1)$:
\begin{align*}
    y = x_1 \cdot \beta + \dots + x_5 \cdot \beta + \epsilon.
\end{align*}

\paragraph*{Scenario 2 (many small effects)}
All variables are drawn from a Bernoulli distribution: $x_j \overset{\text{iid}}{\sim} \mathcal{B}(0.5)$ for $j = 1, \dots, p$.
The outcome is simulated using 50 effect variables, each with an effect size of $\beta = 0.2$, and standard normally distributed noise $\epsilon \sim \mathcal{N}(0,1)$:

\begin{align*}
    y=x_1\cdot\beta + \dots x_{50}\cdot\beta + \epsilon.
\end{align*}

\paragraph*{Scenario 3 (correlated variables)}
Eighty variables are drawn from a Bernoulli distribution: $x_1, \dots, x_{80} \overset{\text{iid}}{\sim} \mathcal{B}(0.5)$.
For each of the variables $x_1, \dots, x_5$, four correlated variables with a correlation of $0.3$ are generated, resulting in the variables $x_{81}, \dots, x_{100}$.
The outcome is simulated using ten effect variables with an effect size of $\beta = 2$ and standard normally distributed noise $\epsilon \sim \mathcal{N}(0,1)$:
\begin{align*}
    y=x_1\cdot\beta + \dots x_{10}\cdot\beta + \epsilon.
\end{align*}

\paragraph*{Scenario 4 (interaction effects)}
All variables are drawn from a Bernoulli distribution: $x_j \overset{\text{iid}}{\sim} \mathcal{B}(0.5)$ for $j = 1, \dots, p$.
The outcome is simulated using five effect variables with an effect size of $\beta = 2$, along with pairwise interactions among ten other variables, and standard normally distributed noise $\epsilon \sim \mathcal{N}(0,1)$:
\begin{align*}
    y=x_1\cdot\beta + \dots x_5\cdot\beta+ x_{6}\cdot x_{7}\cdot\beta + \dots + x_{14}\cdot x_{15}\cdot\beta +    \epsilon.
\end{align*}

\paragraph*{Scenario 5 (binary and continuous variables)}
Five variables are drawn from a standard normal distribution $x_1, \dots, x_5 \overset{\text{iid}}{\sim} N(0,1)$, all others are drawn from a Bernoulli distribution $x_j \overset{\text{iid}}{\sim} \mathcal{B}(0.5)$ for $j = 6, \dots, p$.
The outcome is simulated using five binary and the five continuous effect variables with an effect size of $\beta = 2$, and standard normally distributed noise $\epsilon \sim \mathcal{N}(0,1)$:
\begin{align*}
    y=x_1\cdot\beta + \dots x_{10}\cdot\beta + \epsilon.
\end{align*}

\subsection{Estimands}
For this simulation study we define prediction accuracy, model interpretability, and model stability as estimands of interest. 

\subsection{Methods}\label{subsec:sim:methods}
We compare four different approaches: ARTs with Mondrian CPS, decision trees with Mondrian CPS, multiple ARTs, and multiple decision trees.

\paragraph*{ART with CPS}
A regression RF is trained to predict the continuous outcome $y$ using the R package \emph{ranger} version 0.17.0 \citep{wright_ranger_2017}. The RF hyperparameter \emph{min.node.size} is set to 10\% of the training observations, while all others remain at their default values. Based on this RF, a regression ART with Mondrian CPS is constructed, providing point predictions and 95\% prediction intervals for $y$. In addition, the same ART is used to predict the probability of $y$ exceeding 0.5 ($P(y>0.5)$).
To investigate the influence of the ART hyperparameters, we use $min.bucket = {150,200,250}$, $metric = {SV, WSV, prediction}$, and $probs_quantiles = {\text{no quantiles}, 10\%, 25\%}$.

\paragraph*{Decision tree with CPS}
A regression decision tree with Mondrian CPS is constructed using ranger version 0.17.0 \citep{wright_ranger_2017}, providing point predictions and 95\% prediction intervals for $y$. In addition, this tree is used to predict $P(y>0.5)$. The hyperparameter \emph{min.bucket} is set to the same value as for the ARTs to enable a comparable tree depth.

\paragraph*{Multiple ARTs}
Two separate ARTs are built for each prediction: a regression ART for $y$ and a probability ART for $P(y>0.5)$. For better comparability, the regression ART remains the same as in ARTs with CPS.
For the binary outcome, a probability RF is trained, followed by building a probability ART. The hyperparameters are set analogous to the ART with CPS.

\paragraph*{Multiple decision trees}
Analogous to multiple ARTs, two decision trees are trained. The hyperparameters are set as for decision trees with CPS.

\subsection{Performance measures}\label{subsec:performance_measures}
We evaluate the different ART and decision tree models based on multiple performance measures covering different aspects including interpretability, stability, and prediction accuracy. All performance measures are evaluated across 100 repetitions. 

\paragraph*{Prediction accuracy}
The prediction performance for the continuous outcome $y$ is evaluated using the root mean squared error (RMSE). 
For the evaluation of probability predictions for $P(y>0.5)$, the Brier score is used. For models with CPS, the average width and coverage of the 95\% prediction intervals are calculated across the leaves of each tree.

\paragraph*{Interpretability}
Model complexity was measured by the number of leaves and the length of the deepest path in the tree. For methods that require more than one tree, the maximum value across the trees was used, since the interpretability of the two trees is at least as complex as the highest value.

\paragraph*{Stability}
Stability is assessed by pairwise comparisons of the trees trained in the different scenarios and repetitions. The pairwise distance between the trees of each method is measured in terms of their predictions and split variables, based on the metrics from \citep{banerjee_identifying_2012}. 
The split variable distance measures the number of variables that were used as split variables by both models.
The prediction distance measures the mean squared error (MSE) of two models on a given dataset.
For this purpose, an additional dataset with 1,000 observations is simulated for every scenario and repetition. 

\subsection{Results}
The results are presented for the hyperparameter combination that was subsequently used in the other applications in Sections \ref{sec:benchmark_experiment} and \ref{sec:application_nhanes}. The effect of different ART hyperparameters is examined separately in Section \ref{sec:art_hyperparameter}.
For all methods, results are shown for \emph{min.bucket} = 150, as this was the smallest value that ensured a sufficient number of observations for calibration in all leaves across the simulation study, the benchmark datasets, and the real-data application.
Specifically for ART, no quantiles were used for \emph{probs\_quantiles}, as reducing runtime was not necessary. The \emph{metric} SV was used because this metric resulted in a similar tree depth for ARTs and decision trees, thereby ensuring the highest possible comparability between the methods.

\begin{figure*}[t]
	\includegraphics[width=\textwidth]{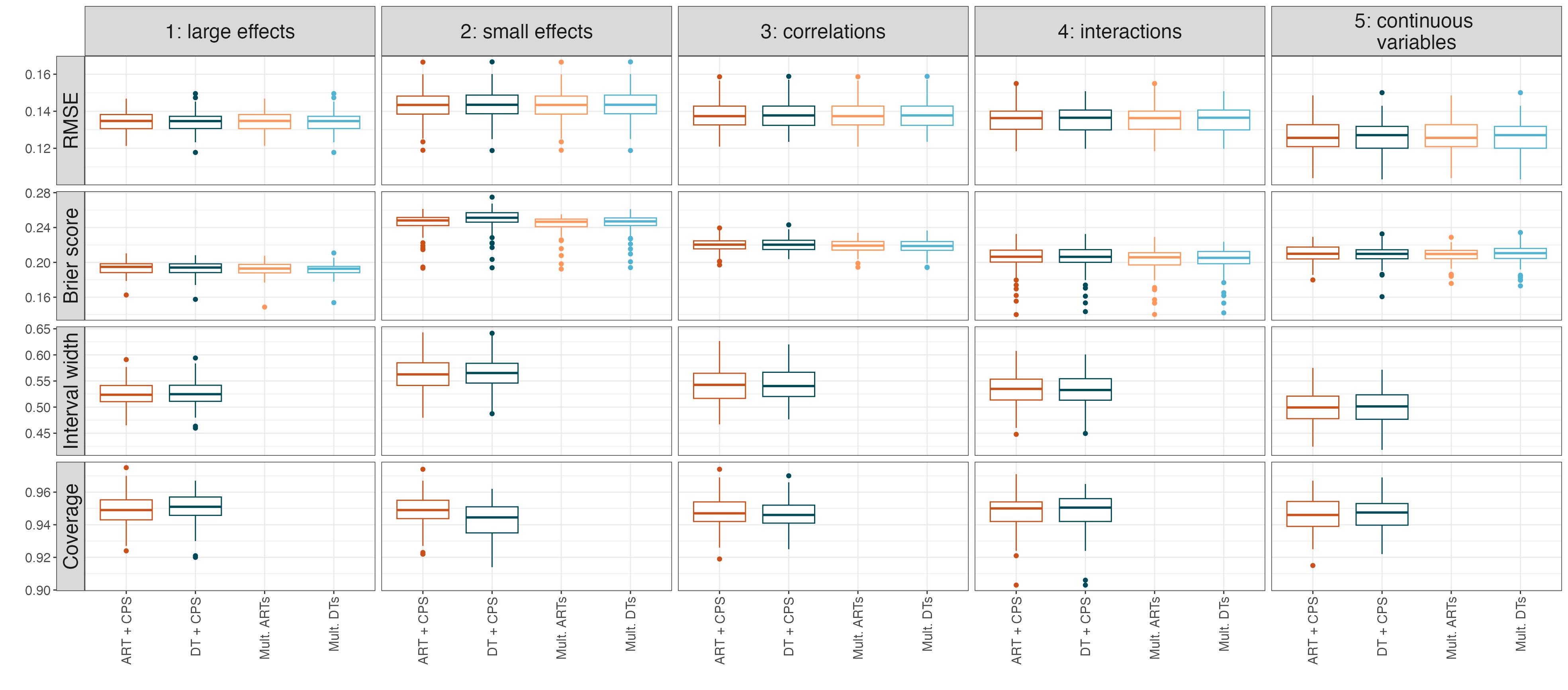}
	\caption{The prediction accuracy measures are shown for the ART with CPS (orange), the decision tree (DT) with CPS (blue), and the multiple ARTs (Mult. ARTs, light orange) and DTs (Mult. DTs light blue) across five simulated scenarios.
	These include the RMSE (A) for predicting the continuous outcome $y$, the Brier score for predicting $P(y>0.5)$, the width of the 95\% prediction interval (C) and the coverage (D) of the prediction interval. For all trees \emph{min.bucket}=150 was used. For ARTs no quantiles and the metric SV were used.
	\label{fig:sim:prediction_accuracy}} 
\end{figure*}

\paragraph*{Prediction accuracy}
For the continuous outcome $y$,  RMSE is very similar for ART with CPS, multiple ARTs, a decision tree with CPS, and multiple decision trees across all repetitions and five simulated scenarios (see Figure \ref{fig:sim:prediction_accuracy}).
The Brier Score for $P(y>0.5)$ shows almost no difference between ART with CPS and the decision tree with CPS. In contrast, in most scenarios the multiple ARTs and multiple decision trees exhibit higher variance in the Brier Score as well as higher overall values (see Figure \ref{fig:sim:prediction_accuracy}).
Regarding the width of the 95\% prediction intervals for the continuous outcome $y$, ART with CPS and decision tree with CPS again display no visible differences across all five scenarios (see Figure \ref{fig:sim:prediction_accuracy}).
Coverage for both ART with CPS and the decision tree with CPS is close to the nominal 95\% prediction interval in all scenarios (see Figure \ref{fig:sim:prediction_accuracy}). In Scenario 2, the coverage of the decision tree with CPS is slightly lower, whereas in Scenario 5 the coverage of ART with CPS is marginally lower.

\paragraph*{Interpretability}
The number of leaves and the maximum tree depth are the same for ART with CPS and the decision tree with CPS in all repetitions and scenarios except scenario 2, in which ART with CPS has one fewer split (see Figure \ref{fig:sim:interpretability_depth_leaves}A). 
Only methods involving CPS are presented, since for multiple ARTs and multiple decision trees the regression tree is the deeper tree and is identical to the respective tree obtained using CPS (see Figure \ref{fig:sim:interpretability_depth_leaves}B).
Therefore, the resulting regression performance is identical.
Overall, interpretability can therefore be considered comparable for this hyperparameter setting.

\begin{figure*}[t]
	\includegraphics[width=\textwidth]{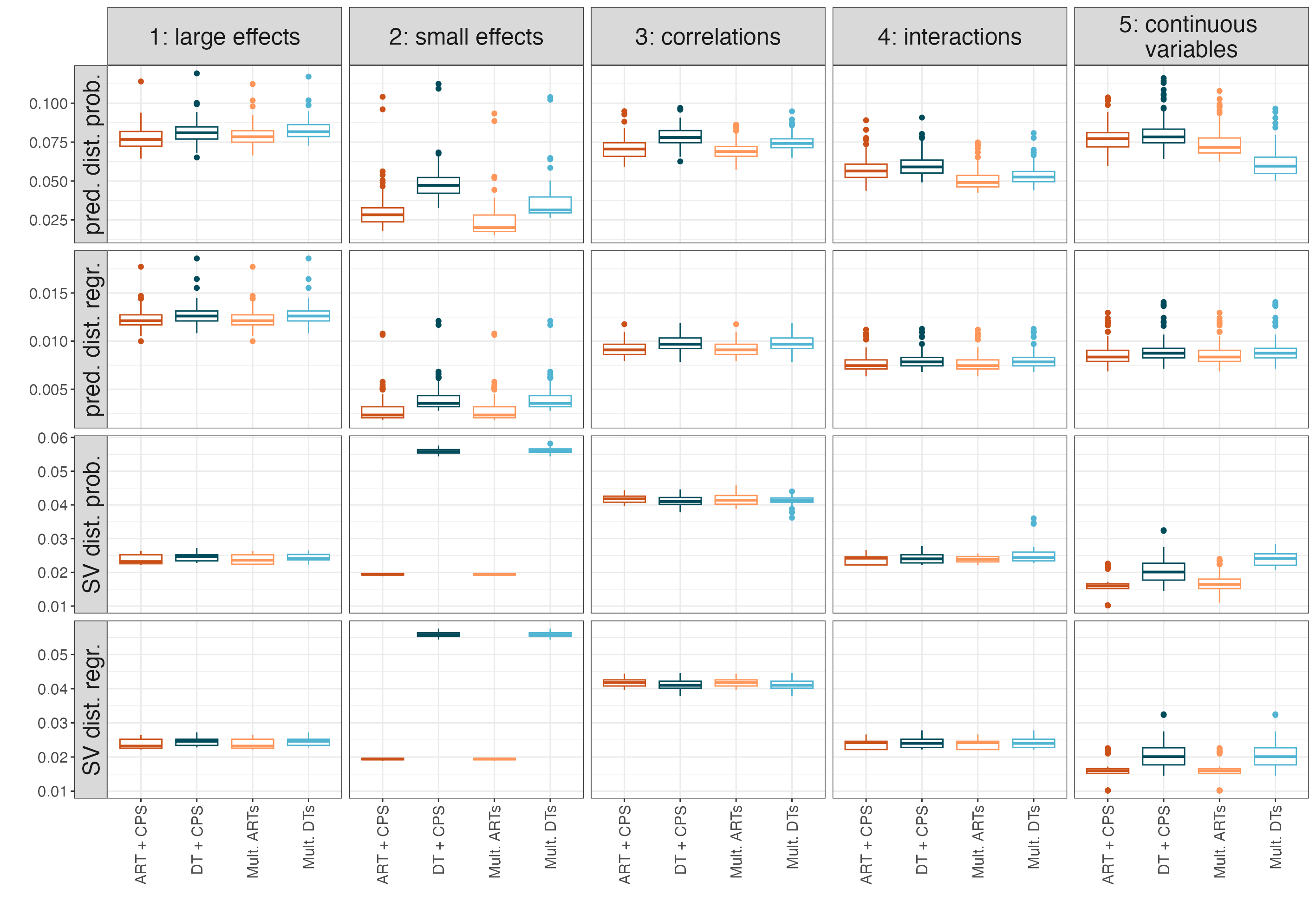}
	\caption{The splitting variable distance (A) and the prediction distance (B) are shown for the ART with CPS (orange), the decision tree (DT) with CPS (blue), and the multiple ARTs (Mult. ARTs, light orange) and DTs (Mult. DTs light blue) across five simulated scenarios. For all trees \emph{min.bucket}=150 was used. For ARTs no quantiles and the metric SV were used.
	\label{fig:sim:stability}} 
\end{figure*}

\paragraph*{Stability}
Across the 100 repetitions, the prediction distance for the predicted probability $P(y>0.5)$ is lowest for ART with CPS, followed by the decision tree with CPS (see Figure \ref{fig:sim:stability}). Except for scenario 2, the methods based on multiple trees exhibit substantially higher prediction distances and are therefore more unstable.
The prediction distance for the regression outcome $y$ is identical for ART with CPS and multiple ARTs, as well as for the decision tree with CPS and multiple decision trees, since in each case the same underlying regression tree is used. Overall, the prediction distance of the ART is smaller than that of the decision tree.

The SV distance for both outcomes for scenario 1,3 and 4 is similar for all methods. ARTs yield a lower distance than decision tree methods in scenarios 2 and 5.

\section{Benchmark experiments}\label{sec:benchmark_experiment}

\subsection{Aim}
To provide a comprehensive performance comparison of ARTs with CPS and other methods,  we also compare the performance of the four methods based on the 21 benchmark dataset used in the publication in which decision trees with conformal prediction were introduced \cite{johansson_interpretable_2018}.

\subsection{Data}
All 21 datasets presented in \cite{johansson_interpretable_2018} are used in this study, summarized in Table \ref{tab:datasets}.
Three of these are from the UC Irvine Machine Learning Repository \footnote{\url{https://archive.ics.uci.edu/}}
, thirteen from the Delve repository \footnote{\url{https://www.cs.toronto.edu/~delve/data/datasets.html}}
 at the University of Toronto, and five from the KEEL repository \footnote{\url{https://sci2s.ugr.es/keel/datasets.php}}
 \cite{alcala-fdez_keel_2010}.
The datasets contain between 1,048 and 9,517 observations with 5 to 15 variables. Except for the abalone dataset, all variables are continuous, as summarized in Table \ref{tab:datasets}.

All datasets are regression problems without missing values. The only preprocessing applied is standardization of the outcomes to the interval [0,1].

For each dataset, the regression outcome is predicted. Additionally, the probability that the outcome exceeds 0.5 (P(y > 0.5)) is estimated in the same way as in the simulation study.
All available variables in each dataset are used as predictors during model training.

\subsection{Methods and performance measures}
Analogous to the simulation study, the four methods ART with Mondrian CPS, decision tree with Mondrian CPS, multiple ARTs, and multiple decision trees are compared as described in Subsection \ref{subsec:sim:methods}. The same hyperparameter settings and variations are used for all methods as in the simulation study. 

For model evaluation, a 10-fold cross-validation is performed. One fold is used as the test dataset. The remaining observations are split into two-thirds for training and one-third for calibration, as described in \citep{johansson_interpretable_2018}.
The same performance measures as in the simulation study are used and are averaged across the ten folds.
The 10-fold cross-validation is repeated 20 times, resulting in 20 averaged performance values per method for evaluation.

\subsection{Results}
For computational reasons, multiple ARTs use 10\% quantiles instead of no quantiles (hyperparameter \emph{probs\_quantiles}). The other hyperparameters are set analogous to the simulation study and the application on NHANES data (Sections \ref{sec:simulation_study} and \ref{sec:application_nhanes}).

\begin{figure*}[t]
	\includegraphics[width=\textwidth]{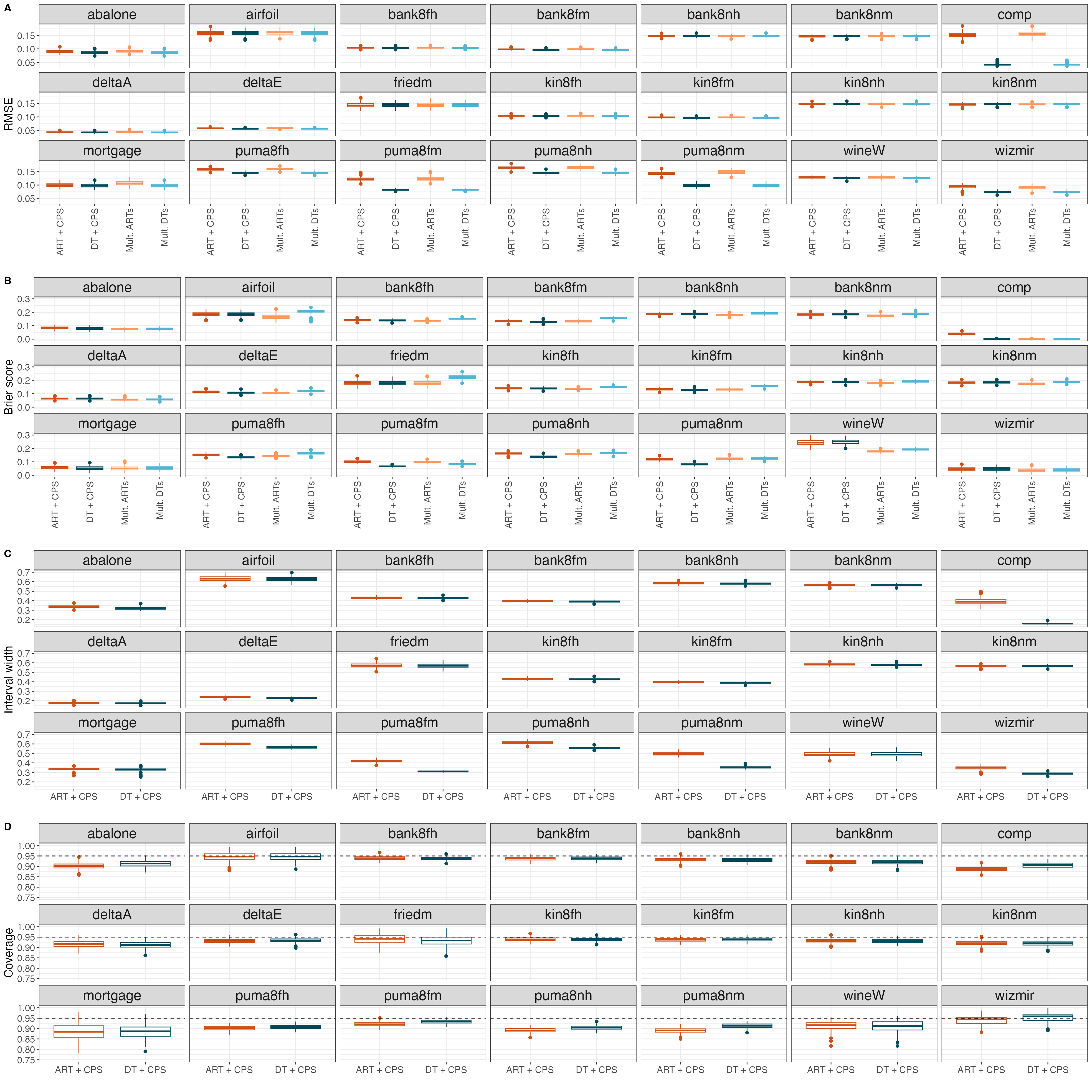}
	\caption{The prediction accuracy measures are shown for the ART with CPS (orange), the decision tree (DT) with CPS (blue), and the multiple ARTs (Mult. ARTs, light orange) and DTs (Mult. DTs light blue) across 21 benchmark datasets.
	These include the RMSE (A) for predicting the continuous outcome $y$, the Brier score for predicting $P(y>0.5)$, the width of the 95\% prediction interval (D) and the coverage (E) of the prediction interval. For all trees \emph{min.bucket}=150 was used. For ART with CPS no quantiles and for multiple ARTs 10\% quantiles were used. Both used the metric SV.
	\label{fig:benchmark:prediction_accuracy}} 
\end{figure*}

\paragraph*{Prediction accuracy}
The RMSE is similar across all methods for most of the 21 benchmark datasets, with a few exceptions where decision trees achieve slightly lower RMSE, for example, in the comp dataset, the datasets from the \emph{puma} family, and \emph{wizmir} (see Figure \ref{fig:benchmark:prediction_accuracy}A).
The Brier Score for the probability that the outcome $y$ exceeds the threshold $P(y>0.5)$ is similar between ART with CPS and decision tree with CPS across all datasets (see Figure \ref{fig:benchmark:prediction_accuracy}B). In contrast, multiple ARTs and multiple decision trees generally exhibit higher Brier Scores and greater variability across nearly all datasets.
Prediction interval widths are also mostly similar between ART and decision tree with CPS. Analogous to the RMSE, decision trees show slightly narrower intervals for individual datasets, such as \emph{comp}, the \emph{puma} family datasets, and \emph{wizmir} (see Figure \ref{fig:benchmark:prediction_accuracy}C). It should be noted that six intervals are unbounded for ARTs with CPS (five in the \emph{airfoil} dataset, one in the \emph{wineW} dataset) and eight for decision trees with CPS (four in the \emph{airfoil} dataset and \emph{wineW} dataset each).
Coverage is similar between ART and decision tree methods. For several datasets, including \emph{mortgage}, \emph{abalone}, \emph{comp}, \emph{deltaA}, the \emph{puma} family, and \emph{wineW}, coverage is slightly below the nominal 95\% (see Figure \ref{fig:benchmark:prediction_accuracy}D).

\begin{figure*}[t]
	\includegraphics[width=\textwidth]{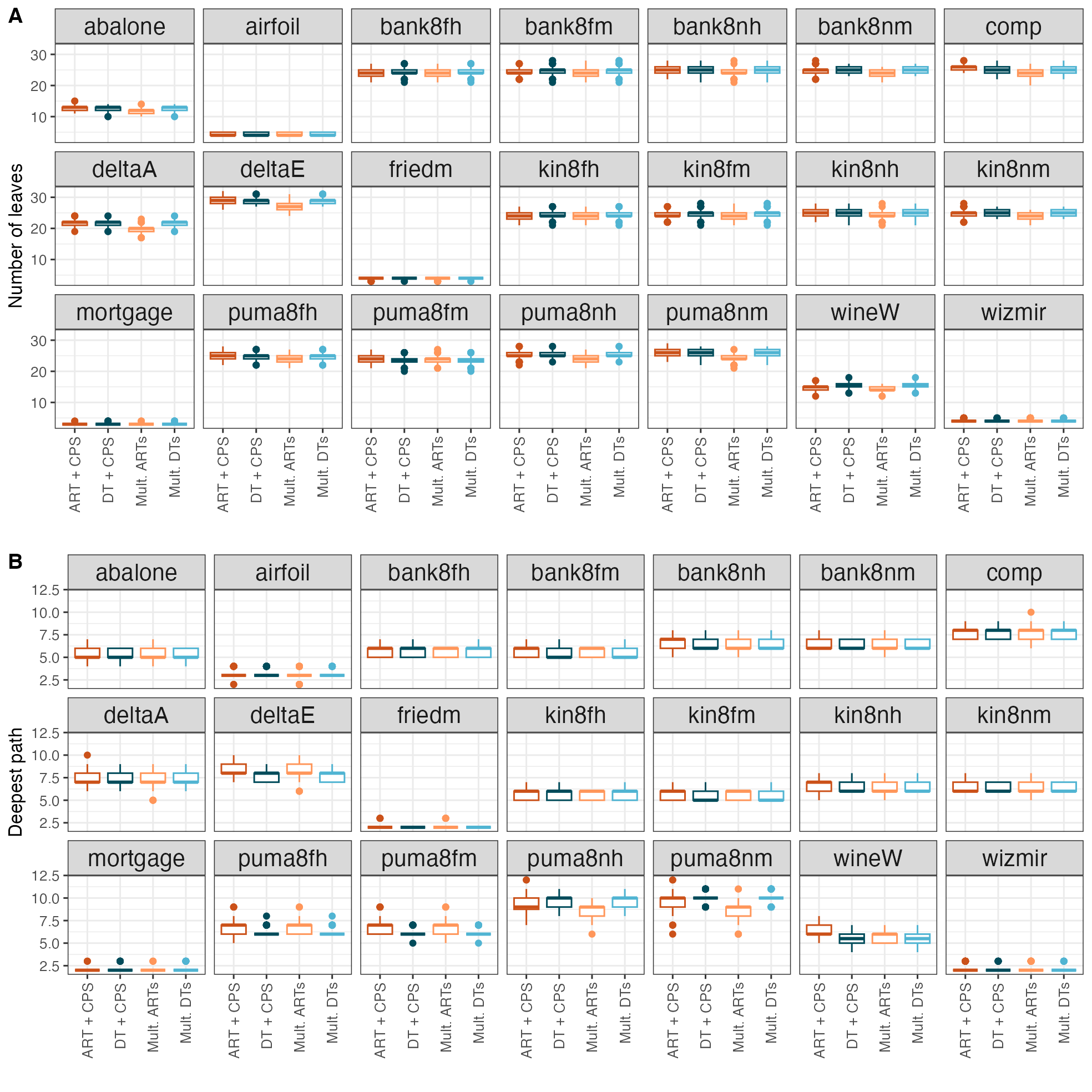}
	\caption{The number of leaves (A) and the deepest path (B) are shown for the ART with CPS (orange), the multiple ARTs (Mult. ARTs, light orange), and the decision tree (DT) with CPS (blue) across 21 benchmark datasets. For all trees \emph{min.bucket}=150 was used. For ART with CPS no quantiles and for multiple ARTs 10\% quantiles were used. Both used the metric SV.
	\label{fig:benchmark:interpretability_depth_leaves}} 
\end{figure*}

\paragraph*{Interpretability}
The number of leaves for ART and decision tree with CPS is almost identical. For multiple ARTs, the number of leaves is sometimes slightly smaller due to restrictions on using only the 10\% quantiles of RF splits (see Figure \ref{fig:benchmark:interpretability_depth_leaves}A). The deepest path in each tree is, on median, of equal depth across all methods (see Figure \ref{fig:benchmark:interpretability_depth_leaves}B).

\begin{figure*}[t]
	\includegraphics[width=\textwidth]{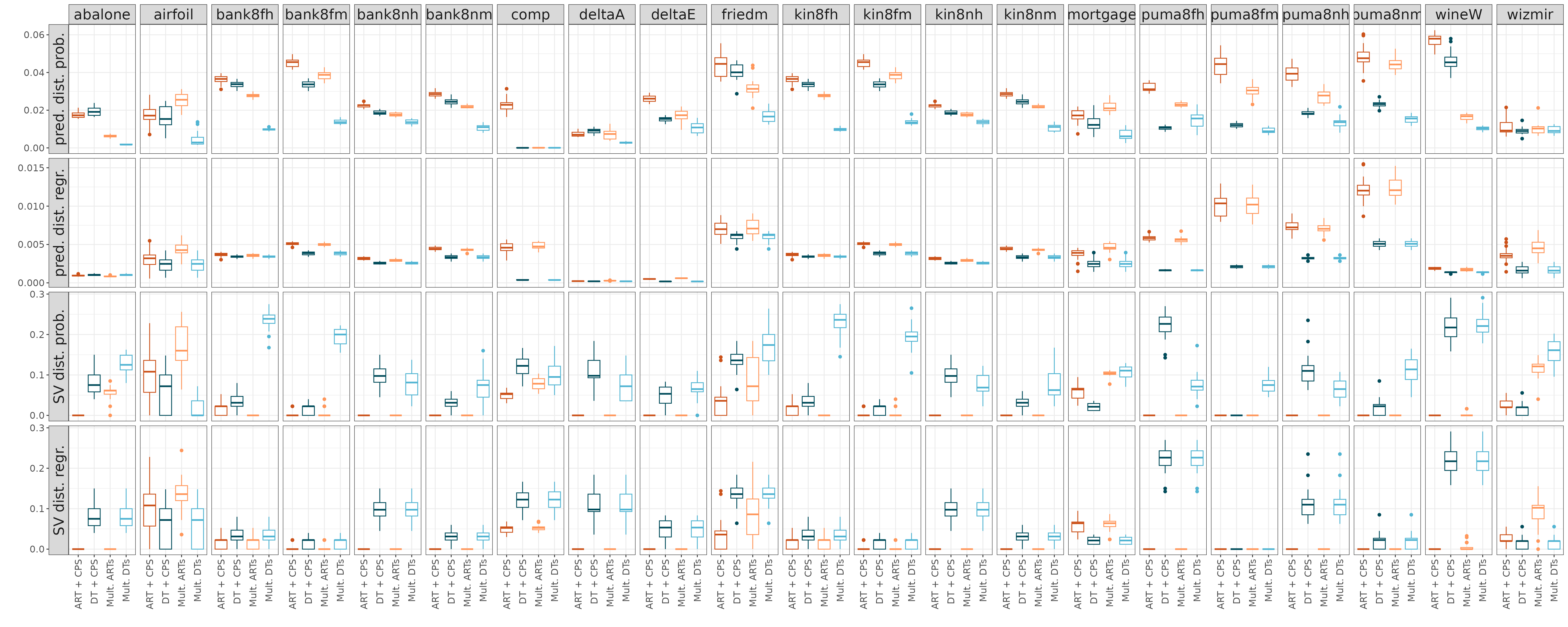}
	\caption{The splitting variable distance (A) and the prediction distance (B) are shown for the ART with CPS (orange), the decision tree (DT) with CPS (blue), and the multiple ARTs (Mult. ARTs, light orange) and DTs (Mult. DTs light blue) across five simulated scenarios. For all trees \emph{min.bucket}=150 was used. For ART with CPS no quantiles and for multiple ARTs 10\% quantiles were used. Both used the metric SV.
	\label{fig:benchmark:stability}} 
\end{figure*}

\paragraph*{Stability}
Across the 20 repetitions of the 10 folds, the prediction distance for the predicted probability $P(y>0.5)$ is lowest for ART with CPS and decision tree with CPS (see Figure \ref{fig:benchmark:stability}), while the highest distance is observed for multiple decision trees.
The prediction distance for the regression outcome $y$ is considerably smaller than that for $P(y>0.5)$ across all methods. Decision tree–based methods exhibit smaller distances than ART-based methods, consistent with the patterns observed in the prediction accuracy results for the \emph{comp}, \emph{puma} family, and \emph{wizmir} datasets.
The split variable distance with respect to both outcomes, the predicted probability $P(y>0.5)$, and the continuous outcome $y$ is lowest for ART with CPS across almost all datasets, followed by multiple ARTs, with the exceptions of \emph{airfoil} and \emph{mortgage}.

\section{Application: diabetes prediction based on NHANES dataset}\label{sec:application_nhanes}

\subsection{Aim}
In this application example, we focus on a medical prediction task, aiming to illustrate how the methods can be applied to a concrete real-world problem.
Specifically, we compare the performance of an ART with CPS, a decision tree with CPS, and ARTs and decision trees in terms of their prediction performance, stability, and interpretability.

\subsection{Data}
We apply our methods to the publicly available NHANES dataset \citep{national_center_for_health_statistics_nhanes_2021}.
NHANES is a nationwide survey conducted by the National Center for Health Statistics (NCHS) to assess the health and nutritional status of adults and children in the United States. The data are publicly available at \url{https://www.cdc.gov/nchs/nhanes/}.
We use the 2017-2020 pre-pandemic cohort \citep{stierman_national_2021} and access the data with the R package \emph{nhanesA} version 1.4.1 \citep{ale_nhanesa_2024}.
The study population consists of participants of all ages, including both adults and children, from the United States. Each year, approximately 5,000 individuals are randomly selected and invited to participate in the study.

\subsection{Data preparation and sample size}
For data preparation, we follow the steps described in the publication \cite{casacchia_development_2024} (see Figure \ref{appendix:fig:flowchart_nhanes} for details). We only deviate from the original workflow by not excluding participants with a diagnosis of prediabetes or diabetes, as this aligns with our research question.
Participants who did not fast or were younger than 18 years are excluded. Individuals taking diabetes medication are also excluded because such treatments may influence glycohemoglobin levels. We removed duplicates and all participants with missing values.
This results in a final dataset of 3,683 participants.

The variables smoking and ancestry are recoded (see Table \ref{tab:nhanes_recoding} for details). The \emph{smoking} variable is recoded into three categories: “never,” “current” and “former.” The \emph{ancestry} variable is grouped into the two most frequent categories (“Non-Hispanic White” and “Non-Hispanic Black”) and a combined “other” category for all remaining participants.
The variable \emph{non\_hdl} is defined as the difference between total cholesterol and HDL. We estimate eGFR using the CKD-EPI Creatinine Equation (2021) from the \emph{kidney.epi} R package version 1.4.0 \citep{boris_kidneyepi_2025}.

\subsection{Outcome}
In addition to the continuous target variable glycohemoglobin, two binary target variables for pre-diabetes and diabetes were defined using thresholds of 5.7\% and  6.5\% glycohemoglobin, respectively, based on the diagnosis criteria for ADA as in \cite{casacchia_development_2024} \citep{american_diabetes_association_professional_practice_committee_2_2023}.

\subsection{Predictors}
We then reduce the dataset to the variables included in the prediction model of the publication \cite{casacchia_development_2024}.
The analysis includes the following six continuous variables: \emph{eGFR}, \emph{non-HDL}, \emph{cholesterol}, \emph{total cholesterol}, \emph{BMI}, \emph{age}, and \emph{glucose}. Additionally, three categorical variables are considered: smoking status (never, former, current), ancestry (Non-Hispanic White, Non-Hispanic Black, Other), and gender (female, male). 

\subsection{Analytical methods and performance measures}
Analogous to the simulation study and the benchmark experiment, the four methods ART with Mondrian CPS, decision tree with Mondrian CPS, multiple ARTs, and multiple decision trees are compared using the hyperparameter settings described in \ref{subsec:sim:methods}. 
The same performance measures as above are used, described in \ref{subsec:performance_measures}.
A 20-times repeated 10-fold cross-validation is performed to obtain point estimates and variability measures for each performance metric.
Two-thirds of the data in each fold are used for training and one-third for calibration. The performance measures are averaged across the ten folds of each repetition.


\subsection{Results}\label{sec:results}

\subsubsection{Prediction accuracy}
\begin{figure*}[t]
	\includegraphics[width=\textwidth]{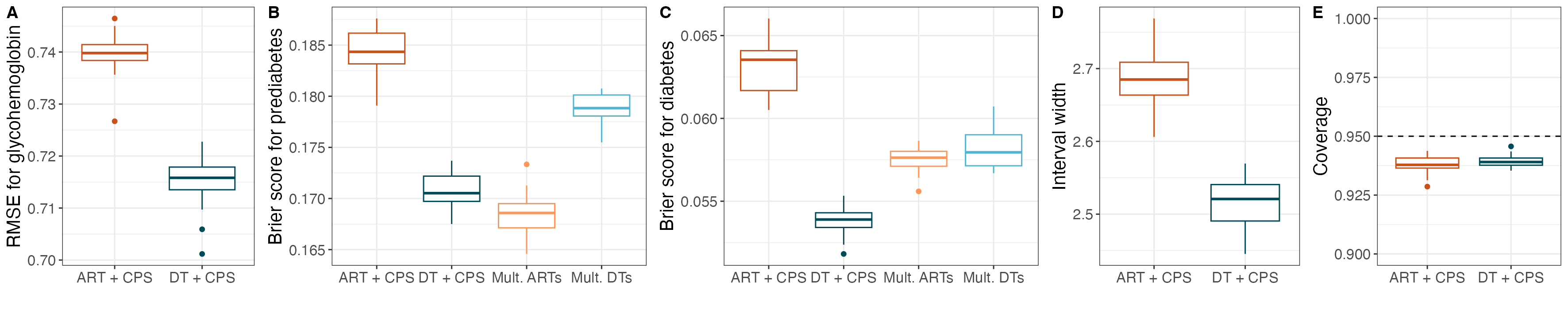}
	\caption{The prediction accuracy measures are shown for the ART with CPS (orange), the decision tree (DT) with CPS (blue), and the multiple ARTs (Mult. ARTs, light orange) and DTs (Mult. DTs light blue) across 20 repetitions.
	These include the RMSE (A) for predicting glycohemoglobin, the Brier score for predicting prediabetes (B) and diabetes (C), the width of the 95\% prediction interval (D) and the coverage (E) of the prediction interval.
	The results are averaged over the 10 folds.
	\label{fig:prediction_accuracy}} 
\end{figure*}

The RMSE for predicting glycohemoglobin is lower for the decision tree models than for the ART models. Because the same ART is used for both ART + CPS and the multiple ARTs, only a single ART result is shown in Figure \ref{fig:prediction_accuracy}A. The same applies to the decision tree models.
For predicting prediabetes and diabetes, the decision tree with CPS again performs best, while the ART with CPS shows slightly worse performance (Figure \ref{fig:prediction_accuracy}B and C). The probability ARTs and decision trees used in the multiple-model approaches perform substantially worse.
The better prediction performance of the decision trees with CPS is also reflected in their narrower prediction intervals, as shown in Figure \ref{fig:prediction_accuracy}D. The coverage of both models with CPS is similar and remains slightly below 95\%, as displayed in Figure \ref{fig:prediction_accuracy}E.

\subsubsection{Interpretability}
\begin{figure*}[t]
	\includegraphics[width=0.5\textwidth]{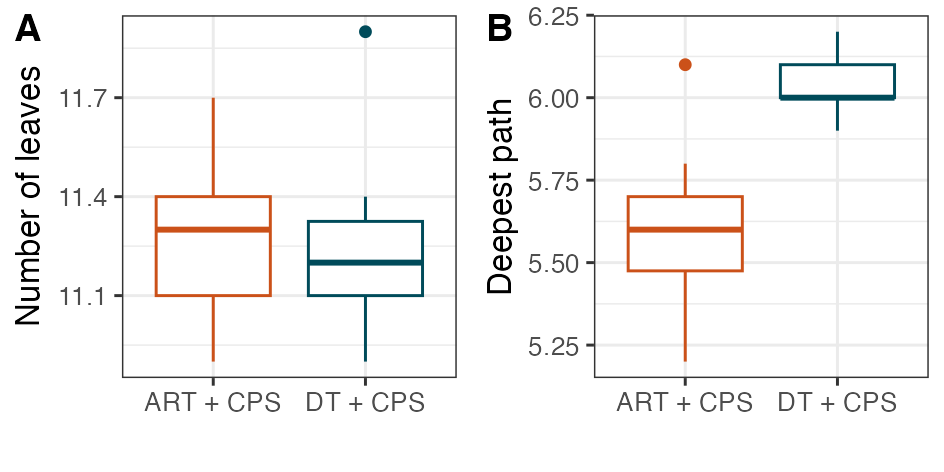}
	\caption{The number of leaves (A) and the deepest path (B) are shown for the ART with CPS (orange) and the decision tree (DT) with CPS (blue) across 20 repetitions. The results are averaged over the 10 folds.
	\label{fig:interpretability_depth_leaves}} 
\end{figure*}

ARTs with CPS and decision trees with CPS have a similar number of leaves and are therefore comparable in overall size, as shown in Figure \ref{fig:interpretability_depth_leaves}. However, ARTs with CPS tend to be wider and less deep, which makes them easier to interpret.
The results for multiple ARTs and multiple decision trees show the same pattern as the trees for predicting glycohemoglobin are the largest.

\subsubsection{Stability}
\begin{figure*}[t]
	\includegraphics[width=0.7\textwidth]{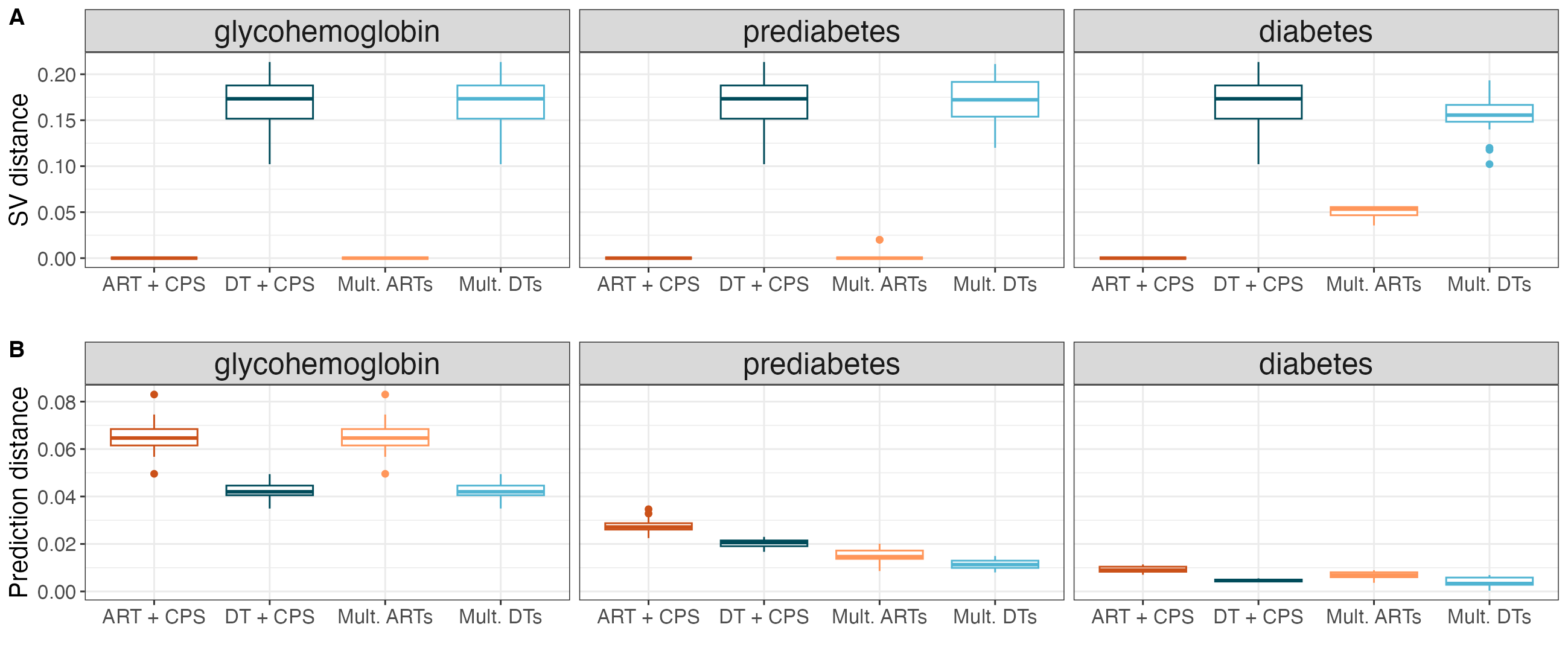}
	\caption{The splitting variable distance (A) and the prediction distance (B) are shown for the ART with CPS (orange), the decision tree (DT) with CPS (blue), and the multiple ARTs (Mult. ARTs, light orange) and DTs (Mult. DTs light blue) across 20 repetitions. The results are averaged over the 10 folds.
	\label{fig:stability}} 
\end{figure*}

ARTs with CPS consistently have the smallest SV distance, close to zero, followed by multiples ARTs. Thus, they use the same split variables and are very stable (see Figure \ref{fig:stability}A). For regression, both ART with CPS and multiple ARTs use the same trees. The same applies to the decision tree-based methods. 
In contrast, decision trees with CPS and multiple decision trees exhibits substantially more variation across folds and repetitions shown by larger distances, leading to reduced interpretability for all predicted outcomes. 

\begin{figure*}[t]
	\includegraphics[width=0.4\textwidth]{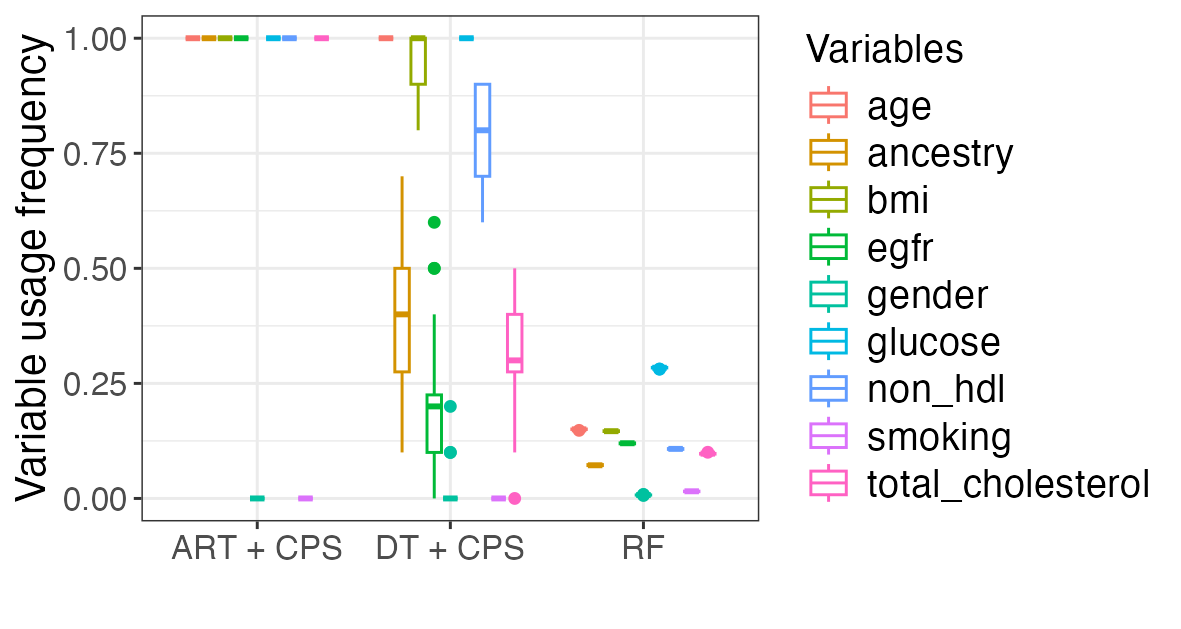}
	\caption{The frequency of variables used as split variables is shown for the ART with CPS, the decision tree (DT) with CPS (blue), and the random forest (RF) across 20 repetitions. The results are averaged over the 10 folds.
	\label{fig:stability_variable_usage}} 
\end{figure*}

To better understand the behavior of ARTs and decision trees with CPS, we examined how often each variable was selected as a split variable across all fitted trees. For comparison, we also evaluated the corresponding split-frequency distribution in the underlying RF, as shown in Figure \ref{fig:stability_variable_usage}. The results indicate that the ART consistently uses exactly the seven variables (age, ancestry, bmi, egfr, glucose, non\_hdl, total\_cholesterol) that the RF most frequently selects for splitting, and each of these variables appears at least once in every ART, reflecting high stability. In contrast, the decision tree always uses the two most frequent RF variables and never uses the two least frequent ones, while the remaining variables show considerable variability in selection across folds and repetitions.
The prediction distance in Figure \ref{fig:stability}B shows that both CPS based methods are similarly stable, with a minimally smaller distance for decision trees with CPS.
The two methods that rely on multiple trees are less stable.

\section{Influence of ART hyperparameters}\label{sec:art_hyperparameter}
We also investigate how ART hyperparameters influence various performance measures in the simulation, the benchmark experiments and the application example. To study the effect on tree depth, we varied the \emph{min.bucket} value from 150 to 250 in increments of 50. For continuous variables, we use selected percentiles of the RF split-point distribution in addition to using all possible split points. Specifically, we use the 25th, 50th and 75th percentiles, as well as the 10th to 90th percentiles in increments of 10 for the hyperparameter \emph{probs\_quantiles}. We also evaluate different similarity \emph{metrics} for distance to the RF, prediction distance, SV and WSV. 
Due to the increased computational time required for multiple ARTs, particularly for benchmark datasets with a large number of observations, only ARTs with runtime acceleration via quantile-based splits were evaluated in terms of the prediction \emph{metric}, in combination with \emph{min.bucket} = 150 or 250.

\subsection*{\emph{min.bucket}}
The hyperparameter \emph{min.bucket} serves as a stopping criterion during tree growth and directly affects the size and depth of the ART, thereby influencing multiple performance measures (see Figures \ref{fig:benchmark:interpretability_depth_leaves}, \ref{appendix:fig:benchmark:hyperparameter_prediction_interpretability_numleaves},\ref{appendix:fig:sim:hyperparameter_interpretability}, and \ref{appendix:fig:interpretability_depth_leaves}).

Higher values of \emph{min.bucket} increase the RMSE and the width of the corresponding prediction intervals in several datasets. This effect is clearly visible in the real-data application using NHANES data (see Figure \ref{appendix:fig:accuracy}A). Among the benchmark datasets, the strongest impact is observed for \emph{comp}, \emph{mortgage}, \emph{puma8fm}, and \emph{wizmir} (see Figure \ref{appendix:fig:benchmark:hyperparameter_prediction_accuracy_rmse}).

In contrast, the effect is less pronounced in the remaining benchmark datasets and in the simulation study (see Figure \ref{appendix:fig:sim:hyperparameter_prediction_accuracy}A). A likely explanation is that, for these datasets, tree growth already stops naturally at higher \emph{min.bucket} values. Consistent with this, the affected datasets show a clear reduction in tree depth and number of leaves as \emph{min.bucket} increases (see Figures \ref{appendix:fig:sim:hyperparameter_interpretability}, \ref{appendix:fig:benchmark:hyperparameter_prediction_interpretability_numleaves}, \ref{appendix:fig:benchmark:hyperparameter_prediction_interpretability_treedepth}, and \ref{fig:interpretability_depth_leaves}).

The Brier score also increased slightly with higher \emph{min.bucket}, analogous to the RMSE. This effect is observable, for example, in the \emph{mortgage} dataset and in the real-data application (see Figures \ref{appendix:fig:benchmark:hyperparameter_prediction_accuracy_brier_score}, \ref{appendix:fig:accuracy}B, and \ref{appendix:fig:accuracy}C).
The impact is less pronounced in the remaining benchmark datasets and in the simulation study (see Figure \ref{appendix:fig:sim:hyperparameter_prediction_accuracy}B).

In contrast, coverage was not affected by \emph{min.bucket}, indicating that the calibration of the uncertainty estimates remained robust across different settings (see Figures \ref{appendix:fig:sim:hyperparameter_prediction_accuracy}, \ref{appendix:fig:benchmark:hyperparameter_prediction_accuracy_coverage}, and \ref{appendix:fig:accuracy}E).

No consistent trend in split variable stability was observed across datasets. Higher \emph{min.bucket} values increased stability for the prediction \emph{metric} on the benchmark datasets \emph{friedm}, \emph{wineW}, and \emph{wizmir}, and in the real-data application without quantiles (see Figures \ref{appendix:fig:benchmark:hyperparameter_prediction_stability_sv} and \ref{fig:stability}). A similar pattern appeared in the simulation study for both SV and WSV \emph{metric} (see Figure \ref{appendix:fig:sim:hyperparameter_stability}).

However, some datasets showed the opposite effect: stability decreased with higher \emph{min.bucket} in the \emph{puma} family, \emph{wineW}, and \emph{abalone} (see Figure \ref{appendix:fig:benchmark:hyperparameter_prediction_stability_sv}).

Regarding prediction stability, higher \emph{min.bucket} reduced stability on the \emph{comp} dataset and in the real-data application, consistent with the observed increase in RMSE (see Figures \ref{appendix:fig:benchmark:hyperparameter_prediction_stability_regression}, \ref{appendix:fig:benchmark:hyperparameter_prediction_stability_probability}, and \ref{fig:stability}). In contrast, on the small \emph{friedm} dataset, higher \emph{min.bucket} improved prediction stability, in line with SV stability. For the remaining datasets and in the simulation study, no clear effect of \emph{min.bucket} on prediction stability was observed (see Figure \ref{appendix:fig:sim:hyperparameter_stability}).

\subsection*{\emph{metric}}
The prediction \emph{metric} generally produced similar results to SV and WSV in terms of RMSE, Brier score, and interval width (see Figures \ref{appendix:fig:sim:hyperparameter_prediction_accuracy}, \ref{appendix:fig:benchmark:hyperparameter_prediction_accuracy_rmse}, \ref{appendix:fig:benchmark:hyperparameter_prediction_accuracy_brier_score},  \ref{appendix:fig:benchmark:hyperparameter_prediction_accuracy_interval_width}, and \ref{appendix:fig:accuracy}).
In some cases, the prediction \emph{metric} yielded slightly lower values.
This was observed, for example, on the \emph{comp} benchmark dataset and in the real-data application.
Coverage was not visibly affected by the choice of \emph{metric} see Figures \ref{appendix:fig:sim:hyperparameter_prediction_accuracy}, \ref{appendix:fig:benchmark:hyperparameter_prediction_accuracy_coverage}, and \ref{appendix:fig:accuracy}).
The number of leaves and the length of the deepest path were similar across all three metrics (see Figures \ref{appendix:fig:sim:hyperparameter_interpretability}, \ref{appendix:fig:benchmark:hyperparameter_prediction_interpretability_numleaves}, \ref{appendix:fig:benchmark:hyperparameter_prediction_interpretability_treedepth}, and \ref{appendix:fig:interpretability_depth_leaves}).
Regarding stability, split variable selection was more stable under SV and WSV than under the prediction \emph{metric} (see Figures \ref{appendix:fig:sim:hyperparameter_stability} and \ref{appendix:fig:benchmark:hyperparameter_prediction_stability_sv}).
In contrast, prediction stability was slightly higher for the prediction \emph{metric} than for SV and WSV (see Figures \ref{appendix:fig:sim:hyperparameter_interpretability}, \ref{appendix:fig:benchmark:hyperparameter_prediction_stability_regression}, \ref{appendix:fig:benchmark:hyperparameter_prediction_stability_probability}, and \ref{appendix:fig:stability}).
This is as expected, since the prediction \emph{metric} focuses on the prediction and the SV and WSV \emph{metric} on the split variables while constructing an ART.

\subsection*{\emph{probs\_quantiles}}
Increasing the restriction on available RF splits by using quantiles (no quantiles vs. 10\% or 25\%) can affect ART growth, as fewer candidate splits are available when quantiles are applied.

Using quantiles (no quantiles vs. 10\% or 25\%) generally leads to higher MSE, Brier score, and wider prediction intervals for several benchmark datasets and in the real data application (see Figures \ref{appendix:fig:benchmark:hyperparameter_prediction_accuracy_rmse}, \ref{appendix:fig:benchmark:hyperparameter_prediction_accuracy_brier_score},  \ref{appendix:fig:benchmark:hyperparameter_prediction_accuracy_interval_width}, and \ref{appendix:fig:accuracy}). The effect is minimal for most datasets, but noticeable for \emph{comp} and \emph{mortgage}. 
For the simulation study, the influence is small, affecting only Scenario 5 with continuous variables (see Figure \ref{appendix:fig:sim:hyperparameter_prediction_accuracy}).
Coverage is unaffected by the use of quantiles, except for the \emph{mortgage} dataset, where coverage is slightly reduced when quantiles are applied (see Figure \ref{appendix:fig:benchmark:hyperparameter_prediction_accuracy_coverage}).
Using quantiles reduces the number of leaves and the depth of ARTs, which is intuitive since fewer splits are available to grow the trees (see Figures \ref{appendix:fig:sim:hyperparameter_interpretability}, \ref{appendix:fig:benchmark:hyperparameter_prediction_interpretability_numleaves}, and \ref{appendix:fig:benchmark:hyperparameter_prediction_interpretability_treedepth}).
For the simulation study and the prediction \emph{metric}, there is no visible effect on the stability (see Figures \ref{appendix:fig:sim:hyperparameter_stability}, \ref{appendix:fig:benchmark:hyperparameter_prediction_stability_sv}, \ref{appendix:fig:benchmark:hyperparameter_prediction_stability_regression}, \ref{appendix:fig:benchmark:hyperparameter_prediction_stability_probability}). However, for the SV and WSV \emph{metric}s, stability varies, with 10\% quantiles yielding slightly higher stability.
No consistent trend is observed for the benchmark data, for example, \emph{airfoil} and \emph{deltaE} exhibit lowest stability at 25\% quantiles, whereas \emph{friedm} and \emph{abalone} behave in the opposite way.
For the real data application, stability is higher when no quantiles are used (see Figure \ref{appendix:fig:stability}).

\section{Concluding discussion}\label{sec:discussion}
We demonstrated that ARTs combined with CPS provide an interpretable and reliable framework for predicting continuous outcomes and probabilities of exceeding predefined thresholds within a single tree. Because these outputs remain attached to the same leaf decision rules, no separate classification tree is required for each threshold.

Across the simulation studies, benchmark datasets, and the application based on the NHANES data, ARTs with CPS achieved prediction accuracy comparable to or better than that of decision trees and multiple tree models, although decision trees performed slightly better in some settings. ARTs often showed superior stability in split variable selection and consistent performance across repetitions. Thus, their main empirical advantage was not uniformly higher predictive accuracy, but improved reproducibility of the selected splitting variables.

The results also illustrate the benefit of deriving threshold probabilities from the predictive distribution of a single continuous-outcome model. In most settings, the CPS-based approaches produced lower and less variable Brier scores than approaches fitting separate regression and probability trees. A single model additionally prevents the continuous and threshold-based predictions from being based on different tree structures. However, these estimates should be interpreted as conformal threshold probabilities under the assumptions of the CPS construction, not automatically as clinically validated diagnostic probabilities.

ART construction balances prediction accuracy and structural stability and is not optimized directly for predictive performance, which may explain why decision trees occasionally achieved slightly better prediction accuracy. 
However, especially in clinical application, the stability and generalizability of prediction models is of much higher value than pure predictive performance. Thus, ARTs can be a good trade-off between predictive performance gained from the underlying RF and stability compared to classical decision trees. 
If predictive accuracy is the primary objective, an ensemble model such as an RF may be preferable, although RF performance was not evaluated in the present comparison. This reflects the general trade-off between predictive performance and the interpretability of a single tree. 

However, the stability and prediction performance of ARTs depend on the chosen hyperparameters. 
For example, larger \emph{min.bucket} values reduce tree depth and the number of leaves but can increase RMSE and prediction interval width, particularly in small datasets such as \emph{mortgage}. The proposed leaf-wise Mondrian CPS further requires a sufficient number of calibration observations within each leaf. Because \emph{min.bucket} controls the number of training rather than calibration observations, calibration counts must be checked separately. When the available calibration sample is insufficient, standard CPS applied to the entire tree may be preferable. This underlines the general need for sufficient training data (especially in the clinical application). With the upcoming application of federated learning approaches, the observed stability of ARTs will be beneficial in this approach of distributed machine learning, where some sites could be used for constructing the ART and other sites for the calibration of the CPS. 

The computational cost of ART construction is generally higher than for decision trees, especially when all RF splits are used, although using quantiles can help mitigate this limitation. 
The choice of hyperparameters should therefore reflect whether prediction accuracy, structural stability, interpretability, or computational efficiency is prioritized. While our systematic hyperparameter investigation provides some guidance, optimal settings may vary across datasets.

The conformal guarantees also require careful interpretation. Under exchangeability, they concern repeated future observations within the prespecified Mondrian categories and do not guarantee conditional coverage for every individual predictor value. Leaf-wise calibration reduces the effective calibration sample and may result in variable or unbounded intervals when individual leaves contain few calibration observations. However, in our simulations and benchmark results, we still observed overall high coverage values close to the nominal 95\% and no substantial outliers below 80\%. 

Although the benchmark datasets can only cover a limited range of sample sizes and predictor types, and the simulation scenarios cannot represent all forms of heteroscedasticity, distribution shift, or model misspecification, we are able to show the properties of ARTs combined with CPS on a wider range of realistic application scenarios. 
Of note, the cross-sectional NHANES analysis was intended to illustrate the joint presentation of predictions, intervals, threshold probabilities, and decision paths. It is not intended as an externally or prospectively validated clinical prediction study and does not establish clinical utility.

Future research should involve the methodological extension of ARTs with CPS, for example by integrating CPS for classification tasks and developing automated hyperparameter optimization.
Further work should also examine which specific types of information provided by ARTs with CPS are most relevant for clinical decision-making and how they influence real-world adoption. 
Interactive visualization tools could help translate the uncertainty-aware outputs of ARTs into more intuitive and actionable interfaces for clinicians and other end users. External and prospective validation would be required before integrating such models into clinical decision support systems.

In summary, combining ARTs with CPS provides a powerful and conceptually unified framework for interpretable, stable, and uncertainty-aware predictions. 
This approach enables a single tree to provide continuous predictions with valid prediction intervals and meaningful diagnostic probabilities. It therefore represents a useful alternative to standard decision trees and multi-model pipelines when transparency and structural stability are important, provided that the assumptions and calibration-data requirements of the CPS construction are met.

\backmatter

\section*{Statements and Declarations}

\subsection*{Data availability}
\begin{enumerate}
  \item The Zenodo repository \url{https://zenodo.org/records/21873965} contains the results from the paper. 
  \item The NHANES data used in this study are publicly available at \url{https://www.cdc.gov/nchs/nhanes/} as described in the manuscript.
\end{enumerate}

\subsection*{Code availability}
\begin{enumerate}
  \item The GitHub repository \url{https://github.com/imbs-hl/ART_uncertainty_paper} contains the full R code, allowing users to reproduce all results and figures. It also includes the code used to preprocess the publicly available NHANES data for the analyses.
  \item The \emph{timbR} package \url{https://github.com/imbs-hl/timbR} provides functions for uncertainty quantification using CPS, which can be applied to ARTs or decision trees in R. It also includes plotting functions for visualizing the results (version 3.3).
\end{enumerate}

\subsection*{Author contributions}

Lea L. Mairhöfer: conceptualization, data curation, formal analysis, investigation,
methodology, software, visualization, writing---original draft,
writing---review \& editing.

Silke Szymczak: conceptualization, supervision, writing---review \& editing.

Björn-Hergen Laabs: conceptualization, funding acquisition, project administration,
supervision, writing---review \& editing.

Tuwe Löfström-Cavallin: conceptualization, funding acquisition, supervision,
writing---review \& editing.

All authors read and approved the final manuscript.

\subsection*{Funding}

This study was funded by the Medical Section of the University of Luebeck
(J01--2024, BL).

Tuwe Löfström-Cavallin acknowledges the Swedish Knowledge Foundation and
industrial partners for financially supporting the research and education
environment on Knowledge Intensive Product Realisation SPARK at Jönköping
University, Sweden. Project: PREMACOP grant no. 20220187.

This work is partially supported by the Ministry of Science and Culture of
Lower Saxony through funds from the program zukunft.niedersachsen of the
Volkswagen Foundation for the ``CAIMed -- Lower Saxony Center for Artificial
Intelligence and Causal Methods in Medicine'' project (grant no. ZN4257).

\subsection*{Competing interests}

The authors have no relevant financial or non-financial interests to disclose.

\subsection*{Ethics approval}
Not applicable to this secondary analysis. The study used publicly available, de-identified NHANES data. 

\subsection*{Consent to participate}
Not applicable.

\subsection*{Consent for publication}
Not applicable.

\bibliography{ART_uncertainty} 

@article{breiman_random_2001,
	title = {Random forests},
	volume = {45},
	issn = {1573-0565},
	language = {en},
	urldate = {2021-05-16},
	journal = {Mach. Learn.},
	author = {Breiman, Leo},
	year = {2001},
	pages = {5--32},
}

@article{wright_ranger_2017,
	title = {ranger: a fast implementation of random forests for high dimensional data in {C}++ and {R}},
	volume = {77},
	copyright = {Copyright (c) 2017 Marvin N. Wright, Andreas Ziegler},
	issn = {1548-7660},
	shorttitle = {ranger},
	language = {en},
	urldate = {2021-05-15},
	journal = {J. Stat. Softw.},
	author = {Wright, Marvin N. and Ziegler, Andreas},
	year = {2017},
	pages = {1--17},
}

@book{molnar_interpretable_2022,
	edition = {2},
	title = {Interpretable {Machine} {Learning}: {A} {Guide} for {Making} {Black} {Box} {Models} {Explainable}},
	url = {https://christophm.github.io/interpretable-ml-book},
	author = {Molnar, Christoph},
	year = {2022},
}

@book{breiman_classification_1984,
	title = {Classification and {Regression} {Trees}},
	isbn = {978-0-412-04841-8},
	publisher = {Taylor \& Francis},
	author = {Breiman, L. and Friedman, J. and Stone, C.J. and Olshen, R.A.},
	year = {1984},
	lccn = {83019708},
}

@inproceedings{laabs_construction_2024,
	address = {Cham},
	title = {Construction of artificial most representative trees by minimizing tree-based distance measures},
	isbn = {978-3-031-63797-1},
	language = {en},
	booktitle = {Explainable {Artificial} {Intelligence}},
	publisher = {Springer Nature Switzerland},
	author = {Laabs, Björn-Hergen and Kronziel, Lea L. and König, Inke R. and Szymczak, Silke},
	editor = {Longo, Luca and Lapuschkin, Sebastian and Seifert, Christin},
	year = {2024},
	pages = {290--310},
}

@inproceedings{johansson_accurate_2014,
	title = {Accurate and interpretable regression trees using oracle coaching},
	booktitle = {2014 {IEEE} {Symposium} on {Computational} {Intelligence} and {Data} {Mining} ({CIDM})},
	author = {Johansson, Ulf and Sönströd, Cecilia and König, Rikard},
	year = {2014},
	pages = {194--201},
}

@article{banerjee_identifying_2012,
	title = {Identifying representative trees from ensembles},
	volume = {31},
	journal = {Stat. Med.},
	author = {Banerjee, Mousumi and Ding, Ying and Noone, Anne-Michelle},
	year = {2012},
	pages = {1601--1616},
}

@article{johansson_interpretable_2018,
	title = {Interpretable regression trees using conformal prediction},
	volume = {97},
	issn = {0957-4174},
	urldate = {2024-12-17},
	journal = {Expert Syst. Appl.},
	author = {Johansson, Ulf and Linusson, Henrik and Löfström, Tuve and Boström, Henrik},
	year = {2018},
	pages = {394--404},
}

@inproceedings{johansson_conformal_2013,
	title = {Conformal {Prediction} {Using} {Decision} {Trees}},
	issn = {2374-8486},
	urldate = {2025-01-19},
	booktitle = {2013 {IEEE} 13th {International} {Conference} on {Data} {Mining}},
	author = {Johansson, Ulf and Boström, Henrik and Löfström, Tuve},
	year = {2013},
	pages = {330--339},
}

@incollection{vovk_algorithmic_2005,
	title = {Algorithmic {Learning} in a {Random} {World}},
	booktitle = {Algorithmic {Learning} in a {Random} {World}},
	author = {Vovk, Vladimir and Gammerman, Alex and Shafer, Glenn},
	year = {2005},
}

@inproceedings{craven_extracting_1995,
	title = {Extracting {Tree}-{Structured} {Representations} of {Trained} {Networks}},
	volume = {8},
	urldate = {2025-02-01},
	booktitle = {Adv {Neural} {Inf} {Process} {Syst} {NeurIPS}},
	publisher = {MIT Press},
	author = {Craven, Mark and Shavlik, Jude},
	year = {1995},
}

@article{johansson_why_2006,
	title = {Why {Not} {Use} an {Oracle} {When} {You} {Got} {One}?},
	volume = {10},
	language = {en},
	journal = {Neural Information Processing},
	author = {Johansson, Ulf and Löfström, Tuve and König, Rikard},
	year = {2006},
}

@article{vovk_nonparametric_2019,
	title = {Nonparametric predictive distributions based on conformal prediction},
	volume = {108},
	issn = {1573-0565},
	language = {en},
	urldate = {2025-04-11},
	journal = {Mach Learn},
	author = {Vovk, Vladimir and Shen, Jieli and Manokhin, Valery and Xie, Min-ge},
	year = {2019},
	pages = {445--474},
}

@article{hullermeier_aleatoric_2021,
	title = {Aleatoric and {Epistemic} {Uncertainty} in {Machine} {Learning}: {An} {Introduction} to {Concepts} and {Methods}},
	volume = {110},
	issn = {0885-6125, 1573-0565},
	shorttitle = {Aleatoric and {Epistemic} {Uncertainty} in {Machine} {Learning}},
	number = {3},
	urldate = {2025-06-24},
	journal = {Mach Learn},
	author = {Hüllermeier, Eyke and Waegeman, Willem},
	month = mar,
	year = {2021},
	note = {arXiv:1910.09457 [cs]},
	pages = {457--506},
}

@inproceedings{lundberg_unified_2017,
	address = {Red Hook, NY, USA},
	series = {{NIPS}'17},
	title = {A unified approach to interpreting model predictions},
	isbn = {978-1-5108-6096-4},
	urldate = {2025-06-25},
	booktitle = {Proceedings of the 31st {International} {Conference} on {Neural} {Information} {Processing} {Systems}},
	publisher = {Curran Associates Inc.},
	author = {Lundberg, Scott M. and Lee, Su-In},
	month = dec,
	year = {2017},
	pages = {4768--4777},
}

@article{guidotti_survey_2018,
	title = {A {Survey} of {Methods} for {Explaining} {Black} {Box} {Models}},
	volume = {51},
	issn = {0360-0300},
	number = {5},
	urldate = {2025-06-25},
	journal = {ACM Comput. Surv.},
	author = {Guidotti, Riccardo and Monreale, Anna and Ruggieri, Salvatore and Turini, Franco and Giannotti, Fosca and Pedreschi, Dino},
	month = aug,
	year = {2018},
	pages = {93:1--93:42},
}

@article{bostrom_explaining_2018,
	title = {Explaining {Random} {Forest} {Predictions} with {Association} {Rules}},
	volume = {5},
	issn = {2363-9881},
	language = {de},
	number = {1},
	urldate = {2025-06-25},
	journal = {Archives of Data Science, Series A (Online First)},
	author = {Boström, Henrik and Gurung, Ram B. and Lindgren, Tony and Johansson, Ulf},
	year = {2018},
	pages = {05},
}

@article{friedman_greedy_2001,
	title = {Greedy function approximation: {A} gradient boosting machine.},
	volume = {29},
	issn = {0090-5364, 2168-8966},
	shorttitle = {Greedy function approximation},
	number = {5},
	urldate = {2025-06-25},
	journal = {Ann. Stat.},
	publisher = {Institute of Mathematical Statistics},
	author = {Friedman, Jerome H.},
	month = oct,
	year = {2001},
	pages = {1189--1232},
}

@inproceedings{bostrom_mondrian_2020,
	title = {Mondrian conformal regressors},
	issn = {2640-3498},
	language = {en},
	urldate = {2025-06-26},
	booktitle = {Proceedings of the {Ninth} {Symposium} on {Conformal} and {Probabilistic} {Prediction} and {Applications}},
	publisher = {PMLR},
	author = {Boström, Henrik and Johansson, Ulf},
	month = aug,
	year = {2020},
	pages = {114--133},
}

@article{casacchia_development_2024,
	title = {Development, validation and recalibration of a prediction model for prediabetes: an {EHR} and {NHANES}-based study},
	volume = {24},
	issn = {1472-6947},
	shorttitle = {Development, validation and recalibration of a prediction model for prediabetes},
	language = {eng},
	number = {1},
	journal = {BMC Med Inform Decis Mak},
	author = {Casacchia, Nicholas J. and Lenoir, Kristin M. and Rigdon, Joseph and Wells, Brian J.},
	month = dec,
	year = {2024},
	pages = {387},
}

@article{alcala-fdez_keel_2010,
	title = {{KEEL} {Data}-{Mining} {Software} {Tool}: {Data} {Set} {Repository}, {Integration} of {Algorithms} and {Experimental} {Analysis} {Framework}},
	volume = {17},
	shorttitle = {{KEEL} {Data}-{Mining} {Software} {Tool}},
	journal = {Journal of Multiple-Valued Logic and Soft Computing},
	author = {Alcala-Fdez, Jesus and Fernández, Alberto and Luengo, Julián and Derrac, J. and Garc'ia, S and Sanchez, Luciano and Herrera, Francisco},
	month = jan,
	year = {2010},
	pages = {255--287},
}

@article{ale_nhanesa_2024,
	title = {{nhanesA}: achieving transparency and reproducibility in {NHANES} research},
	volume = {Apr 15},
	journal = {Database},
	author = {Ale, Laha and Gentleman, Robert and Sonmez, Teresa Filshtein and Sarkar, Deepayan and Endres, Christopher},
	year = {2024},
}

@book{boris_kidneyepi_2025,
	title = {kidney.epi: {Kidney}-{Related} {Functions} for {Clinical} and {Epidemiological} {Research}},
	publisher = {Scientific-Tools.Org},
	author = {Boris, Bikbov},
	year = {2025},
}

@article{american_diabetes_association_professional_practice_committee_2_2023,
	title = {2. {Diagnosis} and {Classification} of {Diabetes}: {Standards} of {Care} in {Diabetes}—2024},
	volume = {47},
	issn = {0149-5992},
	shorttitle = {2. {Diagnosis} and {Classification} of {Diabetes}},
	number = {Supplement\_1},
	urldate = {2026-02-13},
	journal = {Diabetes Care},
	author = {{American Diabetes Association Professional Practice Committee}},
	month = dec,
	year = {2023},
	pages = {S20--S42},
}

@article{stierman_national_2021,
	title = {National {Health} and {Nutrition} {Examination} {Survey} 2017-{March} 2020 {Prepandemic} {Data} {Files}-{Development} of {Files} and {Prevalence} {Estimates} for {Selected} {Health} {Outcomes}},
	issn = {2332-8363},
	language = {eng},
	number = {158},
	journal = {Natl Health Stat Report},
	author = {Stierman, Bryan and Afful, Joseph and Carroll, Margaret D. and Chen, Te-Ching and Davy, Orlando and Fink, Steven and Fryar, Cheryl D. and Gu, Qiuping and Hales, Craig M. and Hughes, Jeffery P. and Ostchega, Yechiam and Storandt, Renee J. and Akinbami, Lara J.},
	month = jun,
	year = {2021},
}

@misc{national_center_for_health_statistics_nhanes_2021,
	title = {{NHANES} 2017–{March} 2020 {Pre}-{Pandemic} {Data} {Release} {Notes}},
	author = {{National Center for Health Statistics}},
	year = {2021},
}

@article{friedrich_role_2024,
	title = {On the role of benchmarking data sets and simulations in method comparison studies},
	volume = {66},
	copyright = {© 2023 The Authors. Biometrical Journal published by Wiley-VCH GmbH.},
	issn = {1521-4036},
	language = {en},
	number = {1},
	urldate = {2026-04-10},
	journal = {Biometrical Journal},
	author = {Friedrich, Sarah and Friede, Tim},
	year = {2024},
	note = {\_eprint: https://onlinelibrary.wiley.com/doi/pdf/10.1002/bimj.202200212},
	pages = {2200212},
}

@inproceedings{bostrom_mondrian_2021,
	series = {Proceedings of {Machine} {Learning} {Research}},
	title = {Mondrian conformal predictive distributions},
	volume = {152},
	booktitle = {Proceedings of the {Tenth} {Symposium} on {Conformal} and {Probabilistic} {Prediction} and {Applications}},
	publisher = {PMLR},
	author = {Boström, Henrik and Johansson, Ulf and Löfström, Tuwe},
	editor = {Carlsson, Lars and Luo, Zhiyuan and Cherubin, Giovanni and An Nguyen, Khuong},
	month = sep,
	year = {2021},
	pages = {24--38},
}

@article{boulesteix_plea_2013,
	title = {A {Plea} for {Neutral} {Comparison} {Studies} in {Computational} {Sciences}},
	volume = {8},
	number = {4},
	journal = {PLOS ONE},
	author = {Boulesteix, Anne-Laure and Lauer, Sabine and Eugster, Manuel J. A.},
	month = apr,
	year = {2013},
	pages = {1--11},
}

@inproceedings{lakshminarayanan_simple_2016,
	title = {Simple and {Scalable} {Predictive} {Uncertainty} {Estimation} using {Deep} {Ensembles}},
	urldate = {2026-07-31},
	author = {Lakshminarayanan, Balaji and Pritzel, A. and Blundell, C.},
	month = dec,
	year = {2016},
}

@article{meinshausen_quantile_2006,
	title = {Quantile {Regression} {Forests}},
	volume = {7},
	issn = {1533-7928},
	number = {35},
	urldate = {2026-07-31},
	journal = {Journal of Machine Learning Research},
	author = {Meinshausen, Nicolai},
	year = {2006},
	pages = {983--999},
}

@article{krzanowski_recursive_2007,
	title = {A recursive partitioning tool for interval prediction},
	volume = {1},
	issn = {1862-5355},
	language = {en},
	number = {3},
	urldate = {2026-09-04},
	journal = {ADAC},
	author = {Krzanowski, Wojtek J. and Hand, David J.},
	month = dec,
	year = {2007},
	pages = {241--254},
}

@article{zhao_interval_2021,
	title = {Interval forecasts based on regression trees for streaming data},
	volume = {15},
	issn = {1862-5355},
	language = {en},
	number = {1},
	urldate = {2026-09-04},
	journal = {Adv Data Anal Classif},
	author = {Zhao, Xin and Barber, Stuart and Taylor, Charles C. and Milan, Zoka},
	month = mar,
	year = {2021},
	pages = {5--36},
}

@article{janitza_computationally_2018,
	title = {A computationally fast variable importance test for random forests for high-dimensional data},
	volume = {12},
	issn = {1862-5347},
	number = {4},
	urldate = {2026-09-04},
	journal = {Adv. Data Anal. Classif.},
	author = {Janitza, Silke and Celik, Ender and Boulesteix, Anne-Laure},
	month = dec,
	year = {2018},
	pages = {885--915},
}


\FloatBarrier
\appendix 
\section*{Appendix}

\begin{figure*}[h!]
	\includegraphics[width=\textwidth]{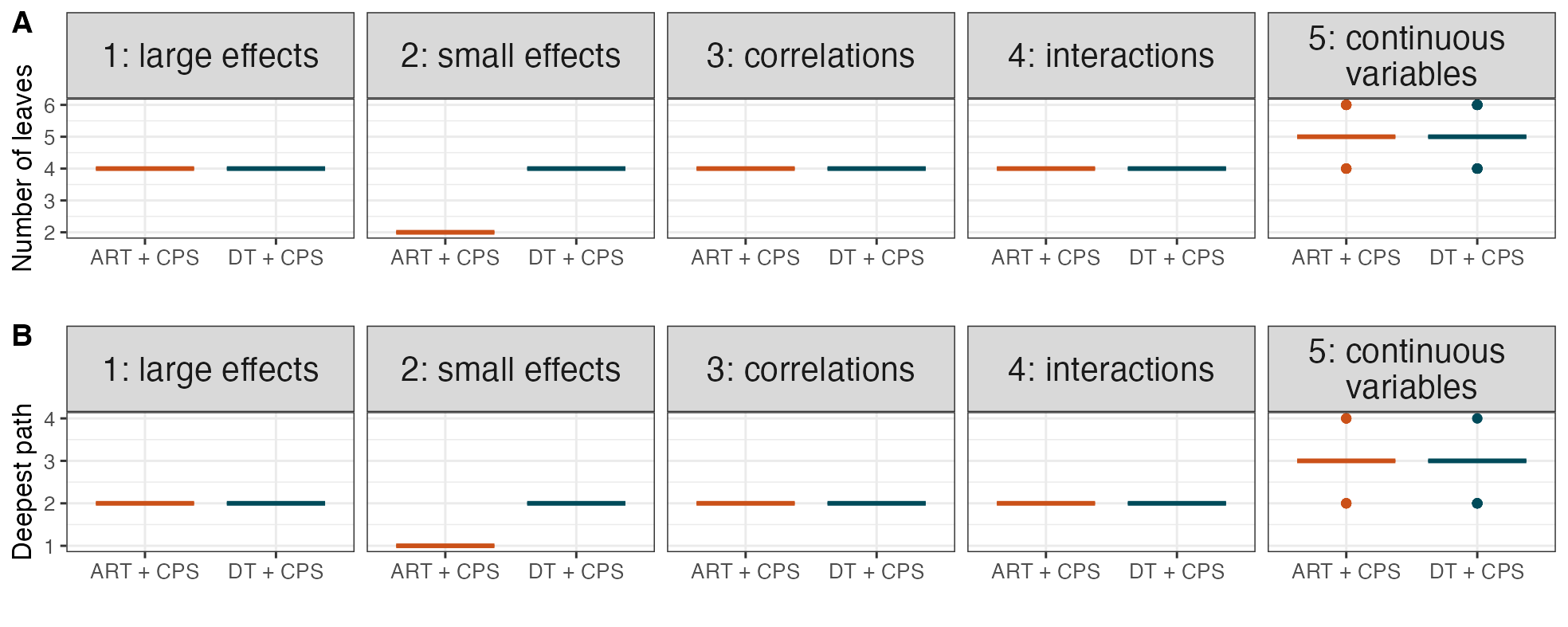}
	\caption{The number of leaves (A) and the deepest path (B) are shown for the ART with CPS (orange) and the decision tree (DT) with CPS (blue) across five simulated scenarios. For all trees \emph{min.bucket}=150 was used. For ARTs no quantiles and the metric SV were used.
	\label{fig:sim:interpretability_depth_leaves}} 
\end{figure*}

\begin{table}[h!]
    \centering
    \caption{Overview of benchmark datasets used to compare ARTs with CPS, decision trees with CPS, multiple ARTs, and multiple decision trees. The name of each dataset, the number of observations $n$, the number of variables $p$, the data repository (source), and the variable types are provided.}
    \label{tab:datasets}
    \scriptsize
    \setlength{\tabcolsep}{3pt}
    \begin{tabular}{l r r l l}
        \hline
        Name & Observations $(n)$ & Variables $(p)$ & Source & Variable type \\
        \hline
        abalone & 4177 & 8 & UCI & 7 continuous, 1 categorical \\
        airfoil & 1503 & 5 & UCI & all continuous \\
        bank8fh & 8192 & 8 & Delve & all continuous \\
        bank8fm & 8192 & 8 & Delve & all continuous \\
        bank8nh & 8192 & 8 & Delve & all continuous \\
        bank8nm & 8192 & 8 & Delve & all continuous \\
        comp & 8192 & 12 & Delve & all continuous \\
        deltaA & 7129 & 5 & KEEL & all continuous \\
        deltaE & 9517 & 6 & KEEL & all continuous \\
       Friedm & 1200 & 5 & KEEL & all continuous \\
        kin8fh & 8192 & 8 & Delve & all continuous \\
        kin8fm & 8192 & 8 & Delve & all continuous \\
        kin8nh & 8192 & 8 & Delve & all continuous \\
        kin8nm & 8192 & 8 & Delve & all continuous \\
        mortgage & 1048 & 15 & KEEL & all continuous \\
        puma8fh & 8192 & 9 & Delve & all continuous \\
        puma8fm & 8192 & 9 & Delve & all continuous \\
        puma8nh & 8192 & 9 & Delve & all continuous \\
        puma8nm & 8192 & 9 & Delve & all continuous \\
        wineW & 3961 & 11 & UCI & all continuous \\
        wizmir & 1460 & 9 & KEEL & all continuous \\
        \hline
    \end{tabular}
\end{table}

\begin{figure*}[h!]
	\includegraphics[width=0.5\textwidth]{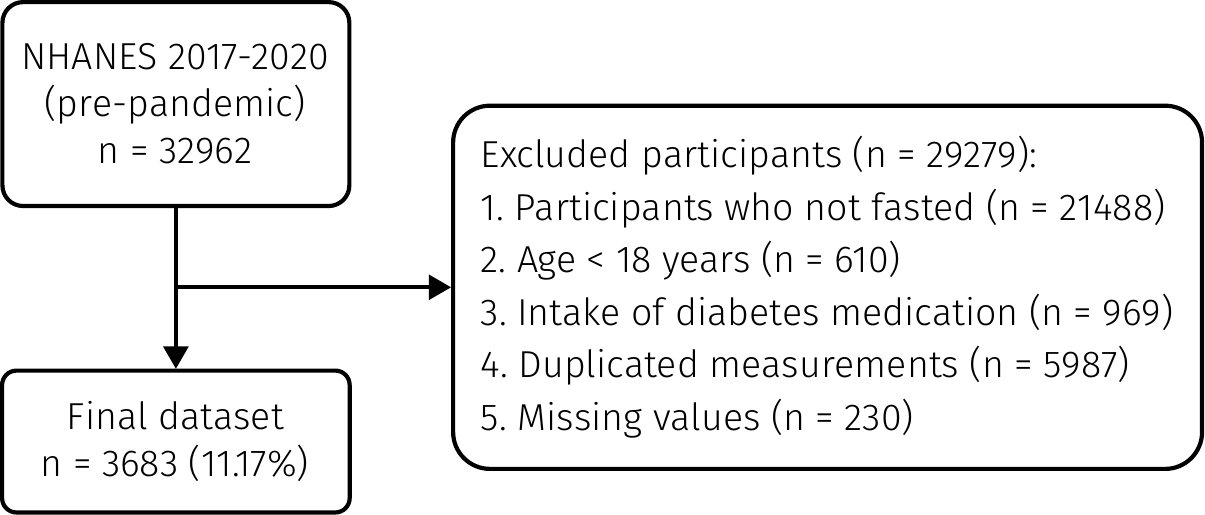}
	\caption{Description of the preprocessing steps for the nhanes dataset. 
	\label{appendix:fig:flowchart_nhanes}} 
\end{figure*}

\begin{table}[h!]
    \centering
    \caption{Original and recoded values of the NHANES variables \emph{smoking} and \emph{ancestry}. Question SMQ020 is about "Smoked at least 100 cigarettes in life", question SMQ040 is about "Do you now smoke cigarettes".}
    \label{tab:nhanes_recoding}
    \begin{tabular}{p{0.15\textwidth}p{0.54\textwidth}p{0.18\textwidth}}
        \hline
        Variable & Original coding & Recoded \\
        \hline
        \multirow{3}{*}{Smoking} & SMQ020 = "No" & never \\
                                 & SMQ020 = "Yes" \& SMQ040 = "Every day"/"Some days" & current \\
                                 & SMQ020 = "Yes" \& SMQ040 = "Not at all" & former \\
        \hline
        \multirow{3}{*}{Ancestry} & Non-Hispanic White & Non-Hispanic White \\
                                   & Non-Hispanic Black & Non-Hispanic Black \\
                                   & Mexican American , Non-Hispanic Asian, Other Hispanic, Other Race - Including Multi-Racial & other \\
        \hline
    \end{tabular}
\end{table}

\begin{figure*}[t]
	\includegraphics[width=\textwidth]{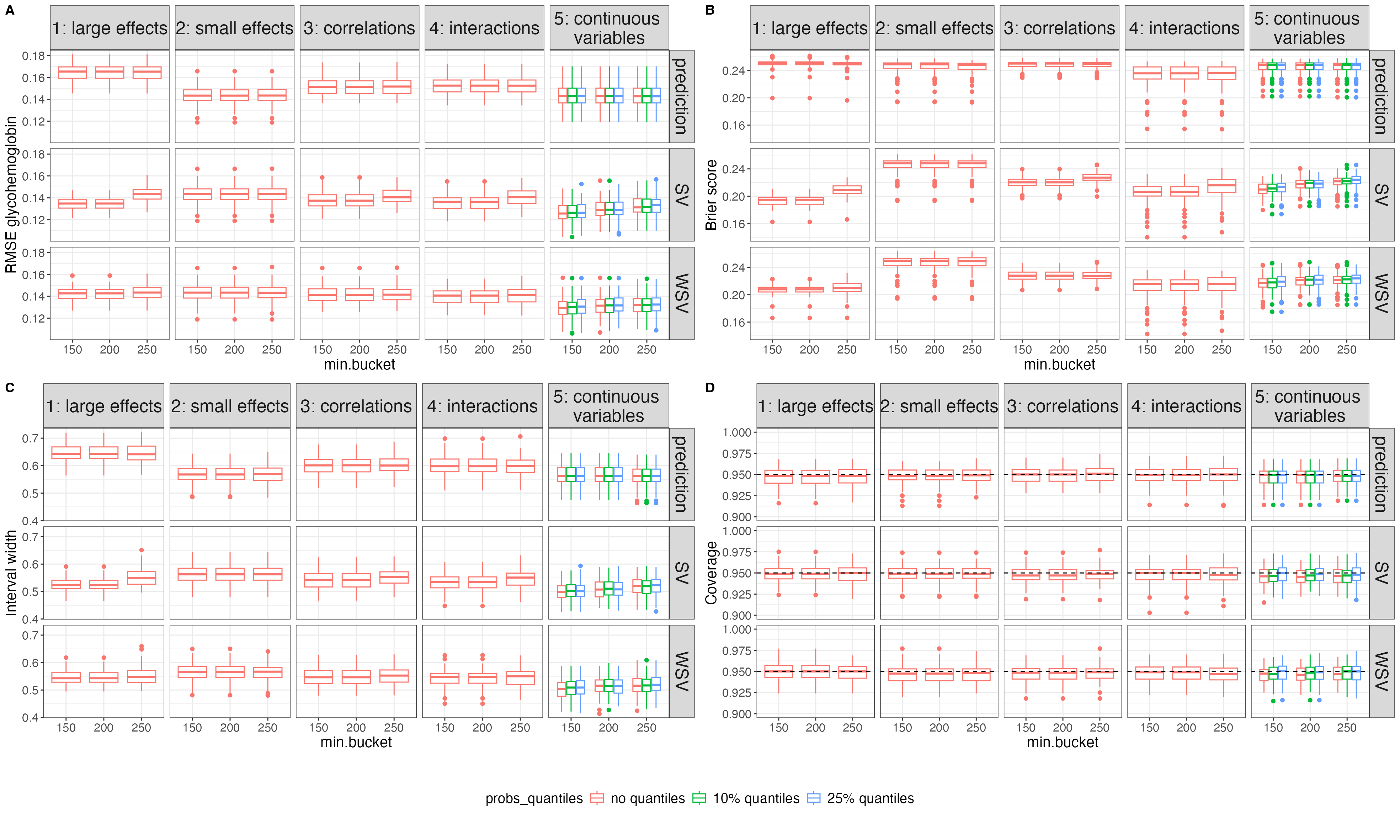}
	\caption{The RMSE (A), the Brier Score (B), the Interval width (C), and the Coverage (D) are shown for the ART with CPS across five simulated scenarios with 100 repetitions. We varied the hyperparameters \emph{min.bucket} (x-axis), \emph{metric} (rows), and \emph{probs\_quantiles} (colors).}
	\label{appendix:fig:sim:hyperparameter_prediction_accuracy}
\end{figure*}

\begin{figure*}[t]
	\includegraphics[width=\textwidth]{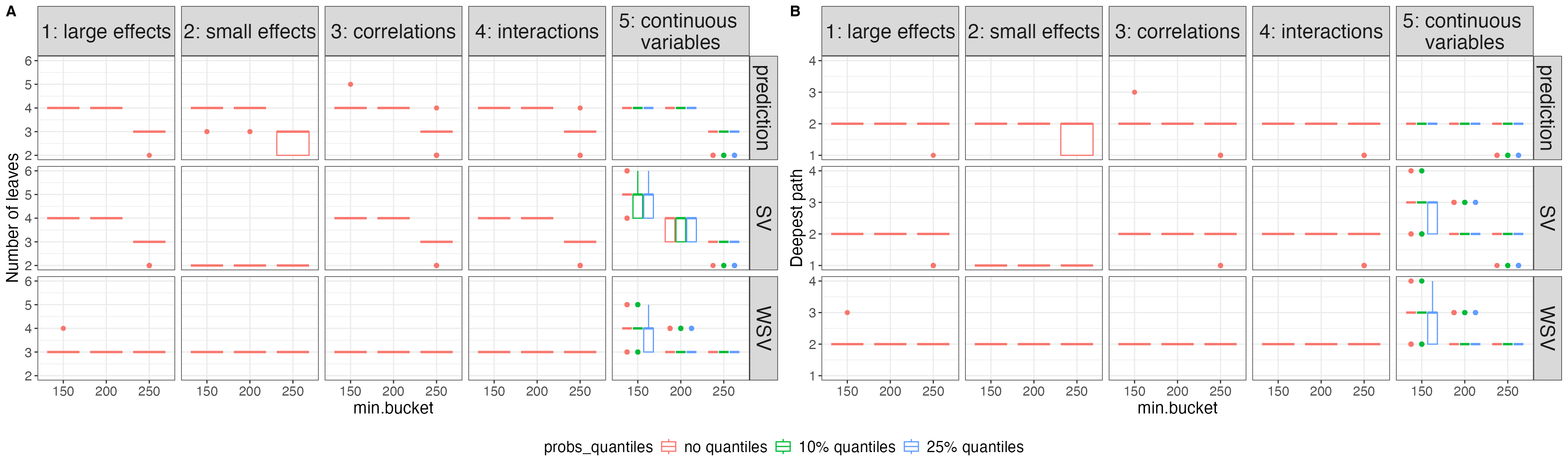}
	\caption{The number of leaves (A) and the deepest path (B) are shown for the ART with CPS across five simulated scenarios with 100 repetitions. We varied the hyperparameters \emph{min.bucket} (x-axis), \emph{metric} (rows), and \emph{probs\_quantiles} (colors).}
	\label{appendix:fig:sim:hyperparameter_interpretability}
\end{figure*}

\begin{figure*}[t]
	\includegraphics[width=0.7\textwidth]{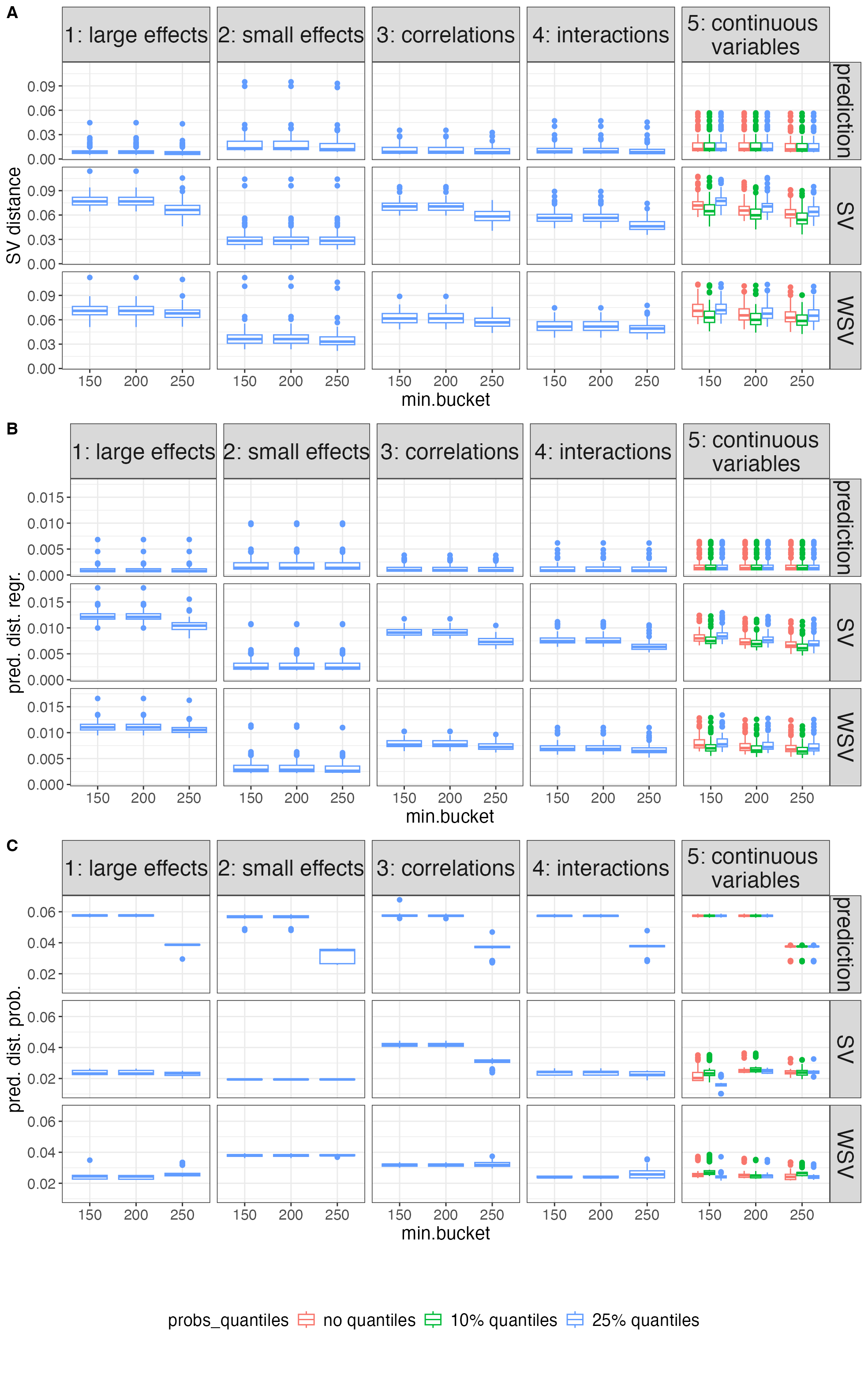}
	\caption{The SV distance (A), the prediction distance for the regression outcome (B) and the prediction distance for the probability outcome (C) are shown for the ART with CPS across five simulated scenarios with 100 repetitions. We varied the hyperparameters \emph{min.bucket} (x-axis), \emph{metric} (rows), and \emph{probs\_quantiles} (colors).}
	\label{appendix:fig:sim:hyperparameter_stability}
\end{figure*}

\begin{figure*}[t]
	\includegraphics[width=0.8\textwidth]{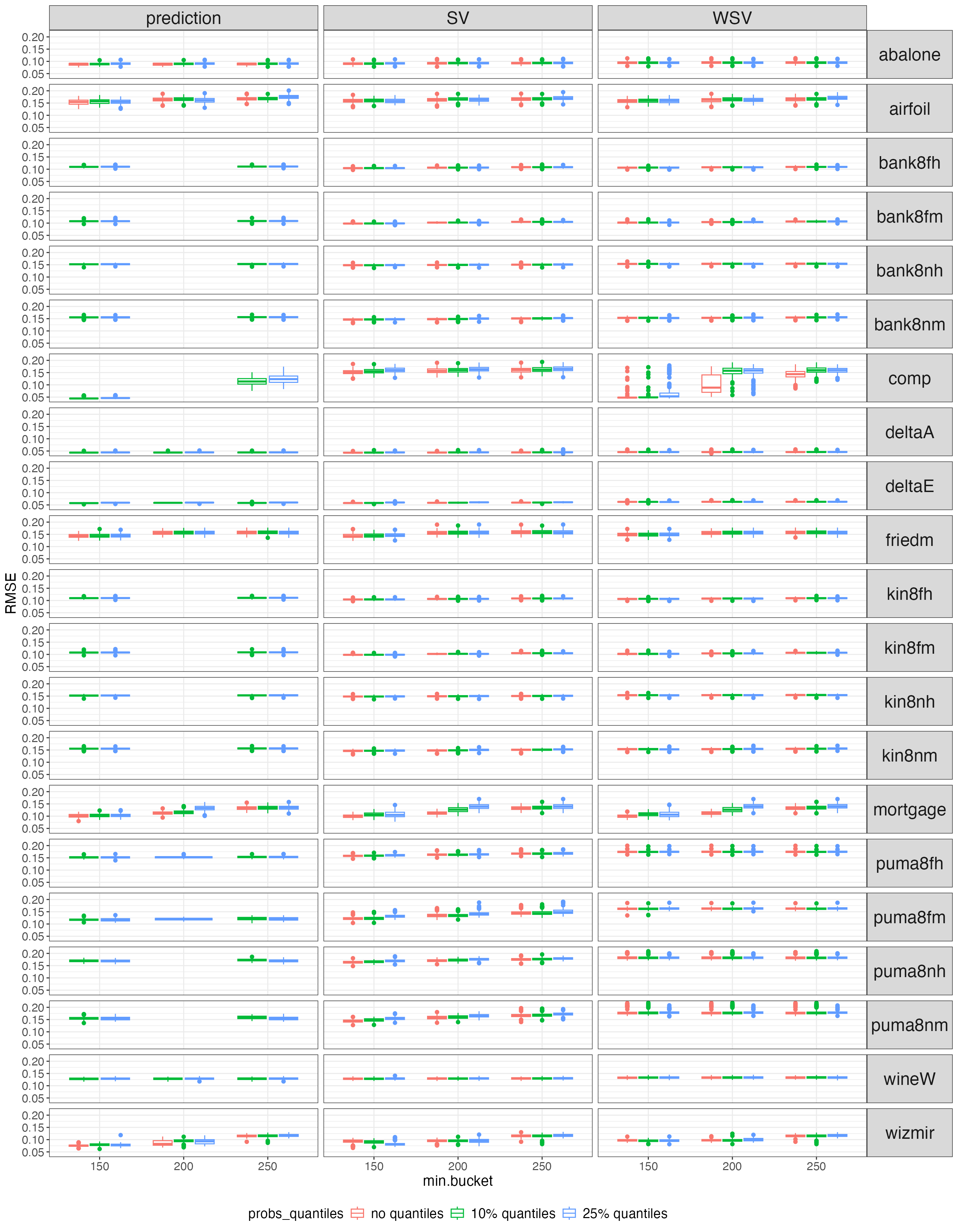}
	\caption{The RMSE is shown for the ART with CPS across 21 benchmark datasets. Each of the 20 repetitions was averaged over 10 folds. We varied the hyperparameters \emph{min.bucket} (x-axis), \emph{metric} (columns), and \emph{probs\_quantiles} (colors).}
	\label{appendix:fig:benchmark:hyperparameter_prediction_accuracy_rmse}
\end{figure*}

\begin{figure*}[t]
	\includegraphics[width=0.8\textwidth]{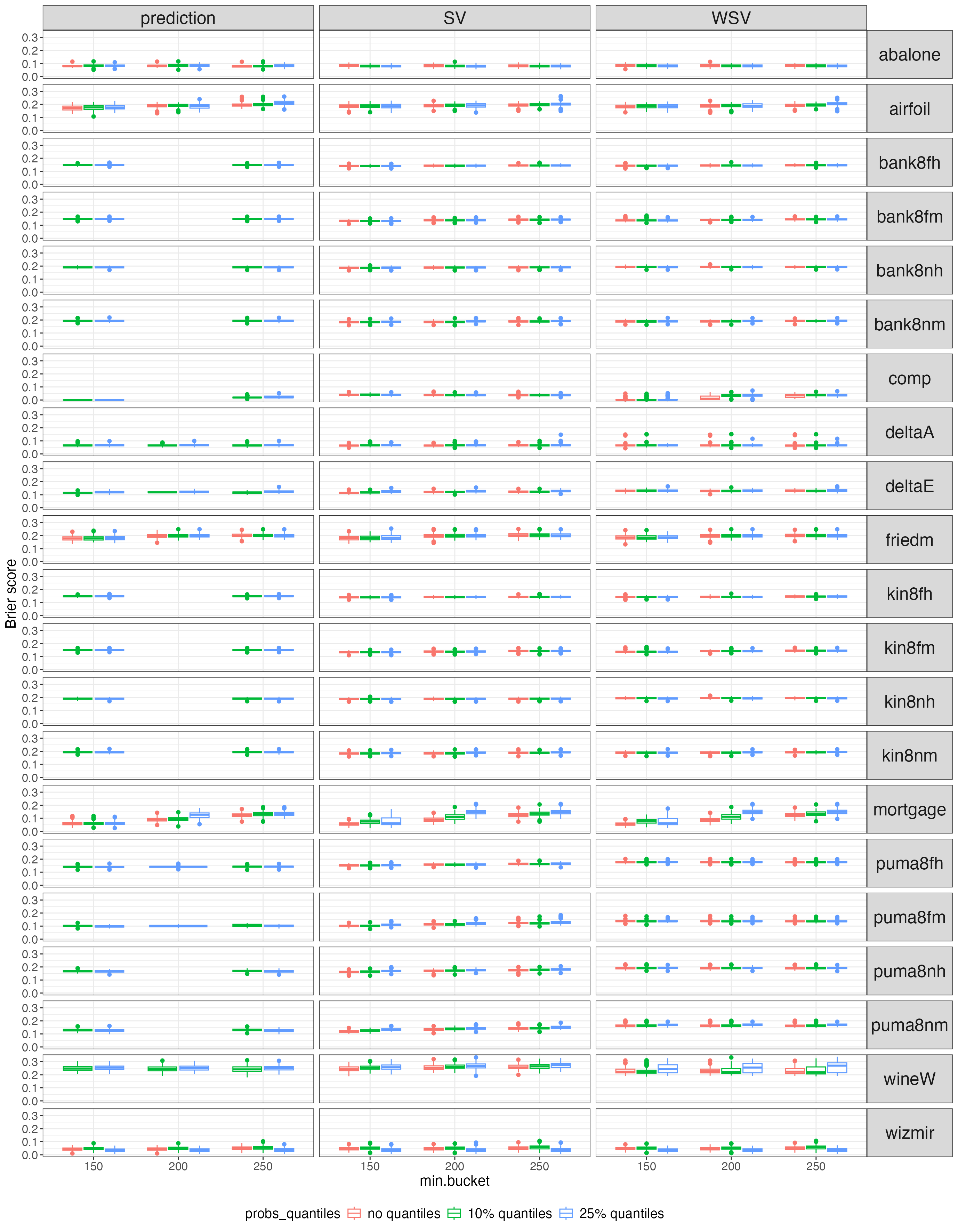}
	\caption{The Brier score is shown for the ART with CPS across 21 benchmark datasets. Each of the 20 repetitions was averaged over 10 folds. We varied the hyperparameters \emph{min.bucket} (x-axis), \emph{metric} (columns), and \emph{probs\_quantiles} (colors).}
	\label{appendix:fig:benchmark:hyperparameter_prediction_accuracy_brier_score}
\end{figure*}

\begin{figure*}[t]
	\includegraphics[width=0.8\textwidth]{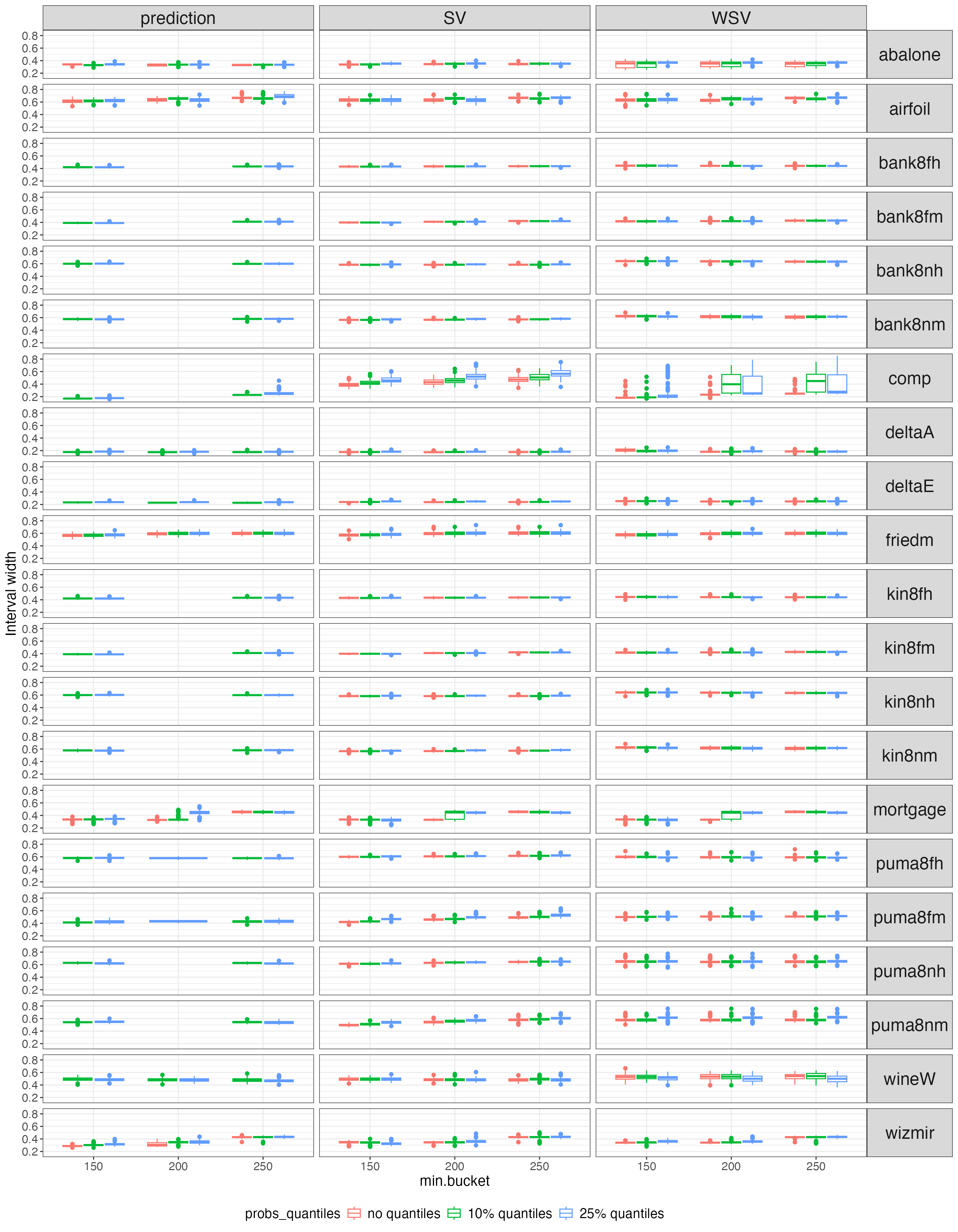}
	\caption{The prediction interval width is shown for the ART with CPS across 21 benchmark datasets. Each of the 20 repetitions was averaged over 10 folds. We varied the hyperparameters \emph{min.bucket} (x-axis), \emph{metric} (columns), and \emph{probs\_quantiles} (colors).}
	\label{appendix:fig:benchmark:hyperparameter_prediction_accuracy_interval_width}
\end{figure*}

\begin{figure*}[t]
	\includegraphics[width=0.8\textwidth]{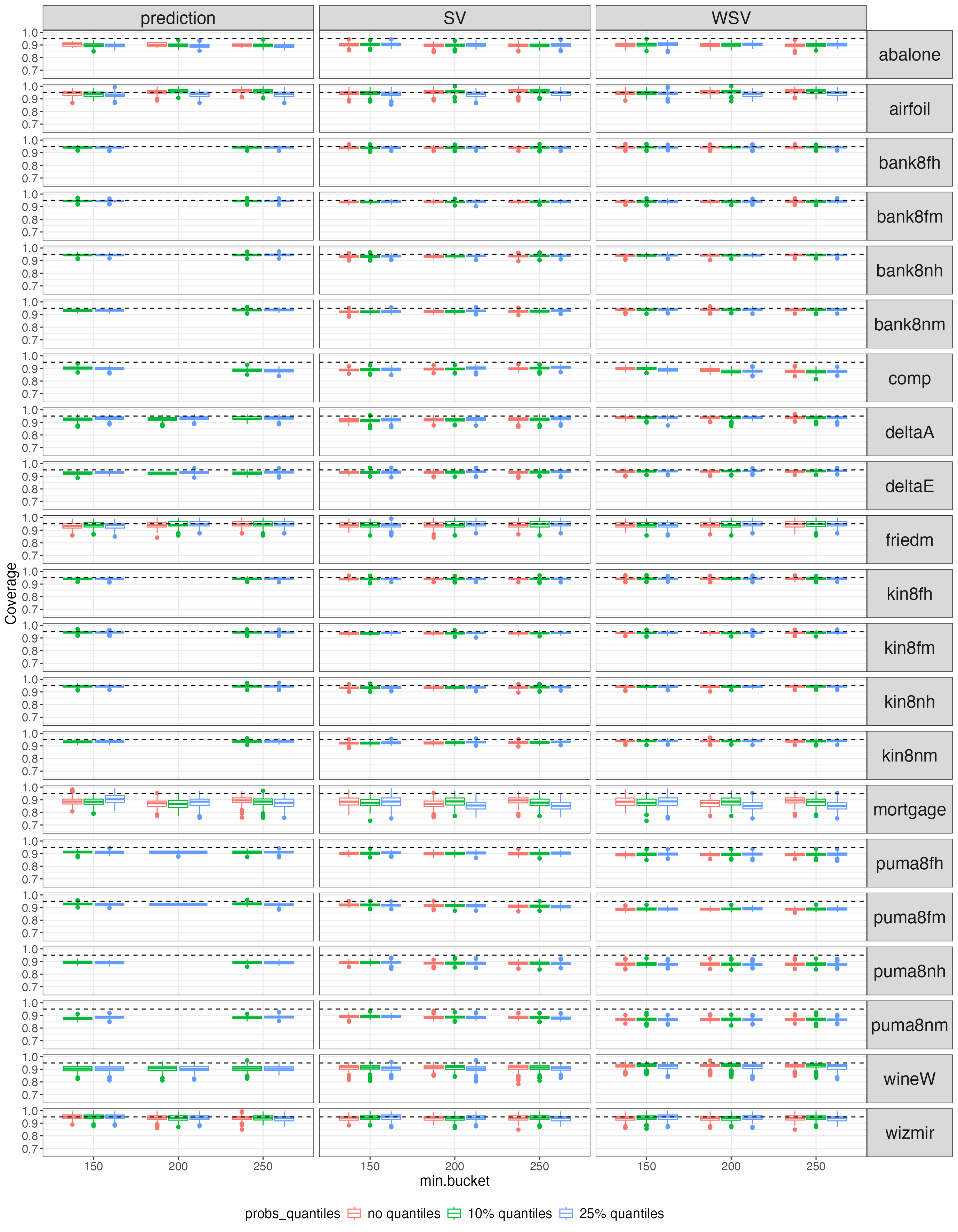}
	\caption{The coverage is shown for the ART with CPS across 21 benchmark datasets. Each of the 20 repetitions was averaged over 10 folds. We varied the hyperparameters \emph{min.bucket} (x-axis), \emph{metric} (columns), and \emph{probs\_quantiles} (colors).}
	\label{appendix:fig:benchmark:hyperparameter_prediction_accuracy_coverage}
\end{figure*}

\begin{figure*}[t]
	\includegraphics[width=0.8\textwidth]{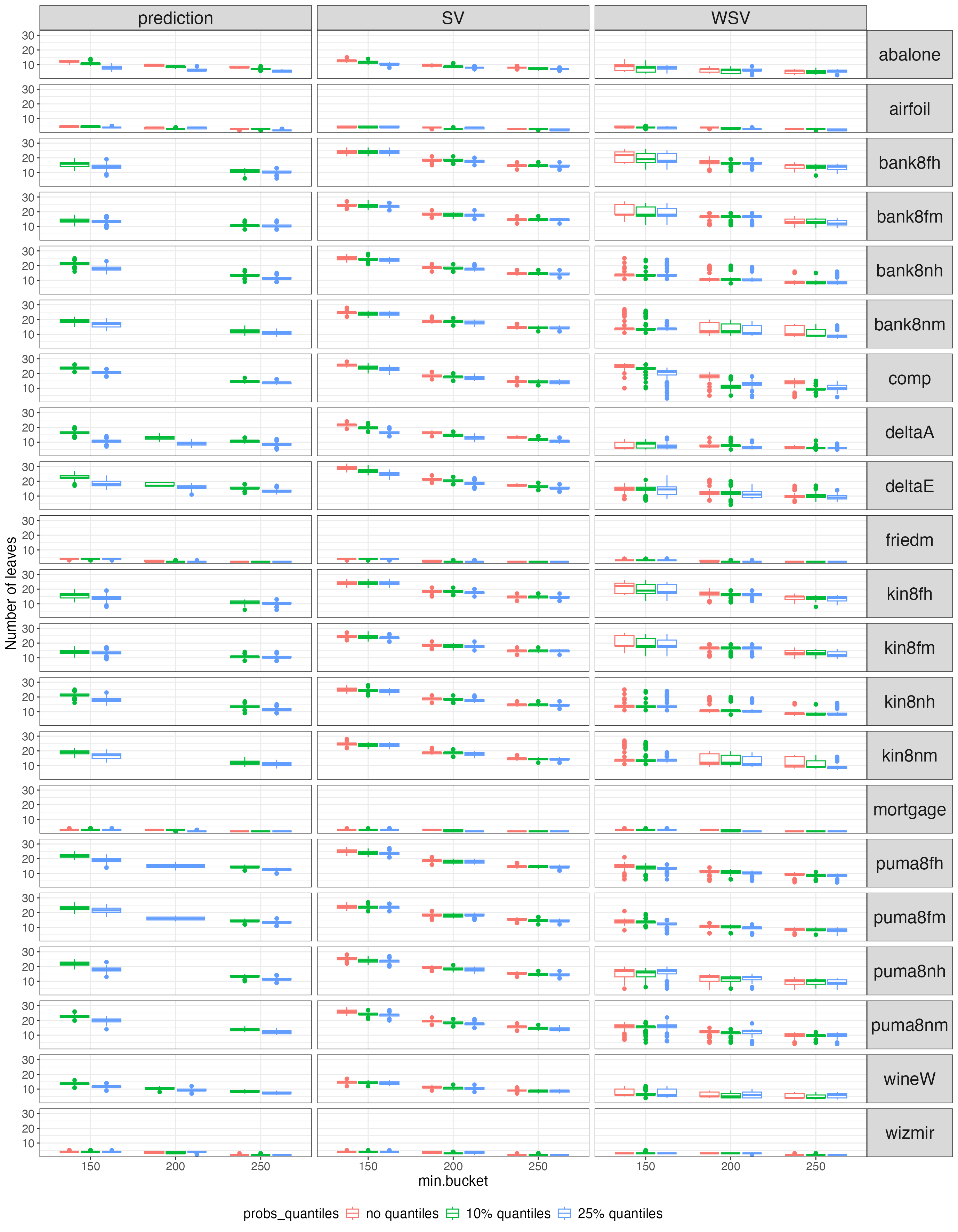}
	\caption{The number of leaves is shown for the ART with CPS across 21 benchmark datasets. Each of the 20 repetitions was averaged over 10 folds. We varied the hyperparameters \emph{min.bucket} (x-axis), \emph{metric} (columns), and \emph{probs\_quantiles} (colors).}
	\label{appendix:fig:benchmark:hyperparameter_prediction_interpretability_numleaves}
\end{figure*}

\begin{figure*}[t]
	\includegraphics[width=0.8\textwidth]{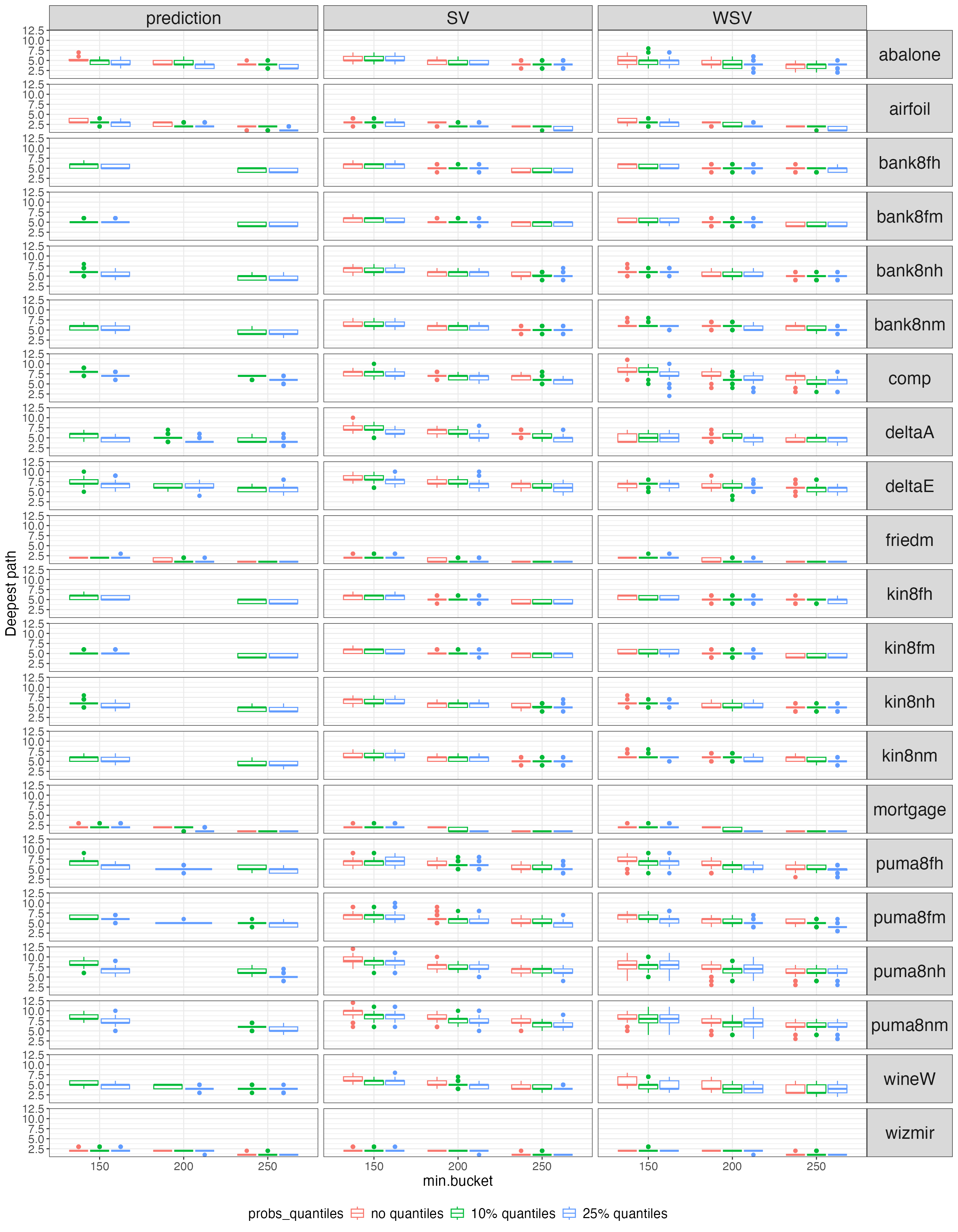}
	\caption{The deepest path is shown for the ART with CPS across 21 benchmark datasets. Each of the 20 repetitions was averaged over 10 folds. We varied the hyperparameters \emph{min.bucket} (x-axis), \emph{metric} (columns), and \emph{probs\_quantiles} (colors).}
	\label{appendix:fig:benchmark:hyperparameter_prediction_interpretability_treedepth}
\end{figure*}

\begin{figure*}[t]
	\includegraphics[width=0.8\textwidth]{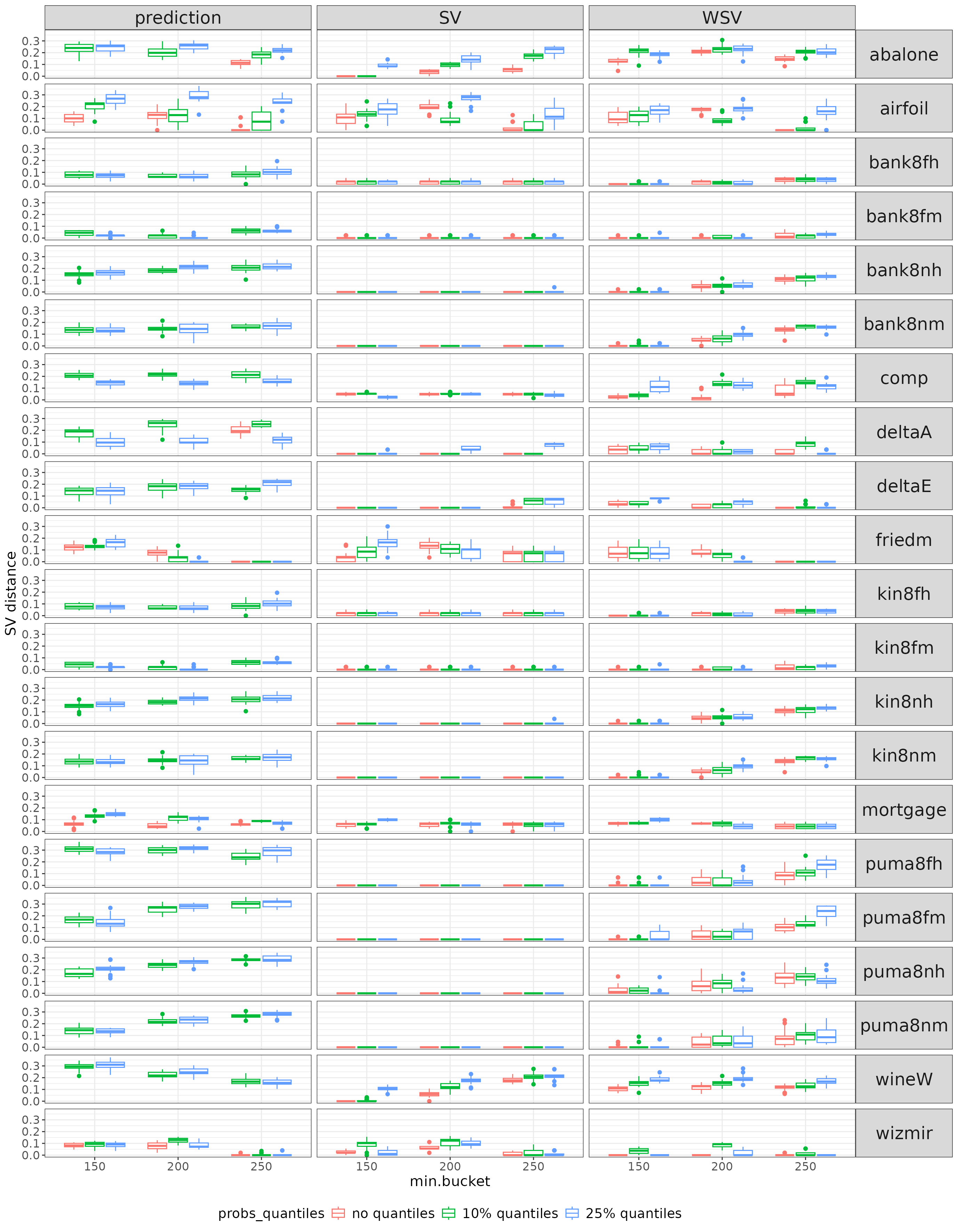}
	\caption{The SV distance is shown for the ART with CPS across 21 benchmark datasets. Each of the 20 repetitions was averaged over 10 folds. We varied the hyperparameters \emph{min.bucket} (x-axis), \emph{metric} (columns), and \emph{probs\_quantiles} (colors).}
	\label{appendix:fig:benchmark:hyperparameter_prediction_stability_sv}
\end{figure*}

\begin{figure*}[t]
	\includegraphics[width=0.8\textwidth]{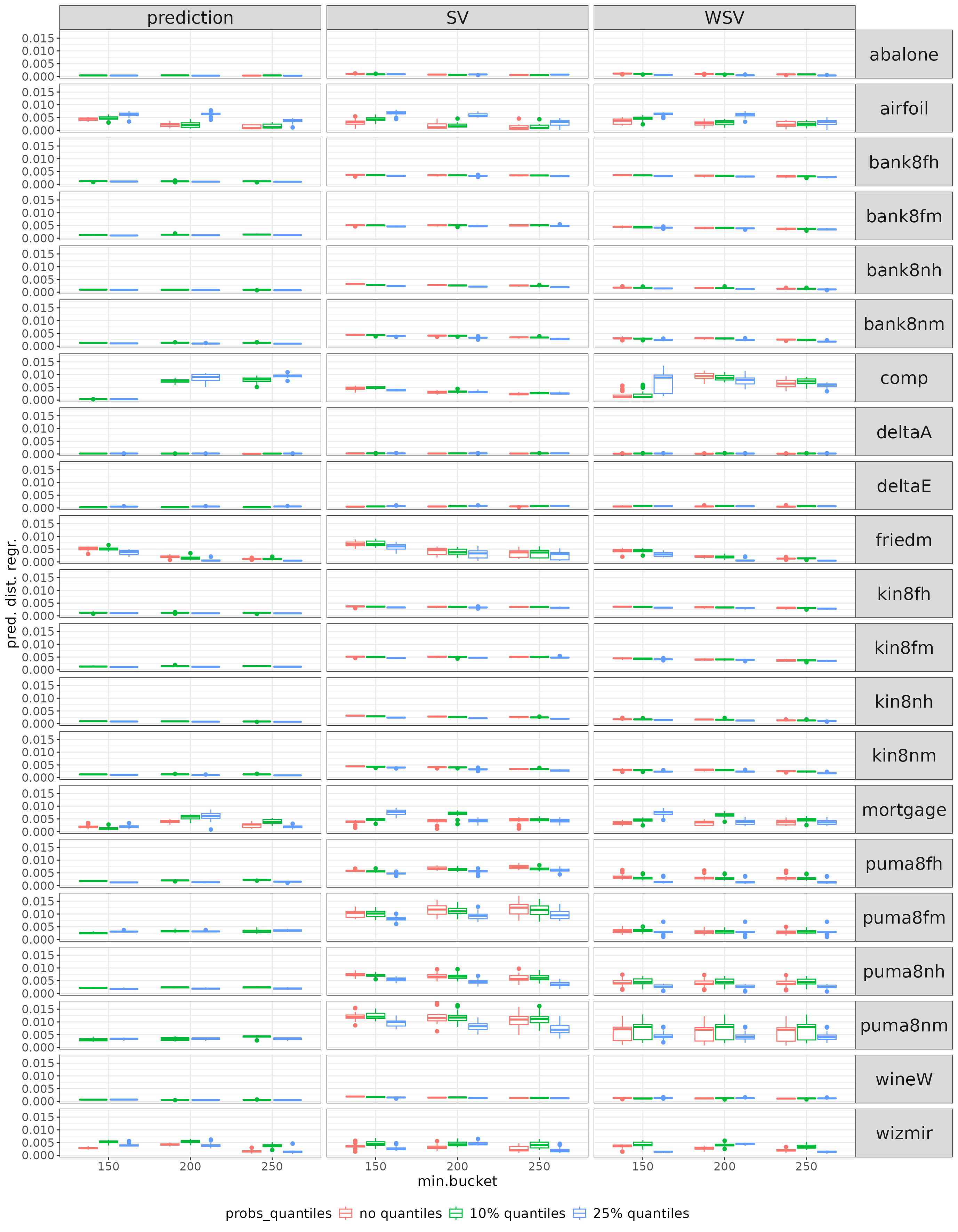}
	\caption{The prediction distance for the continuous outcome is shown for the ART with CPS across 21 benchmark datasets. Each of the 20 repetitions was averaged over 10 folds. We varied the hyperparameters \emph{min.bucket} (x-axis), \emph{metric} (columns), and \emph{probs\_quantiles} (colors).}
	\label{appendix:fig:benchmark:hyperparameter_prediction_stability_regression}
\end{figure*}

\begin{figure*}[t]
	\includegraphics[width=0.8\textwidth]{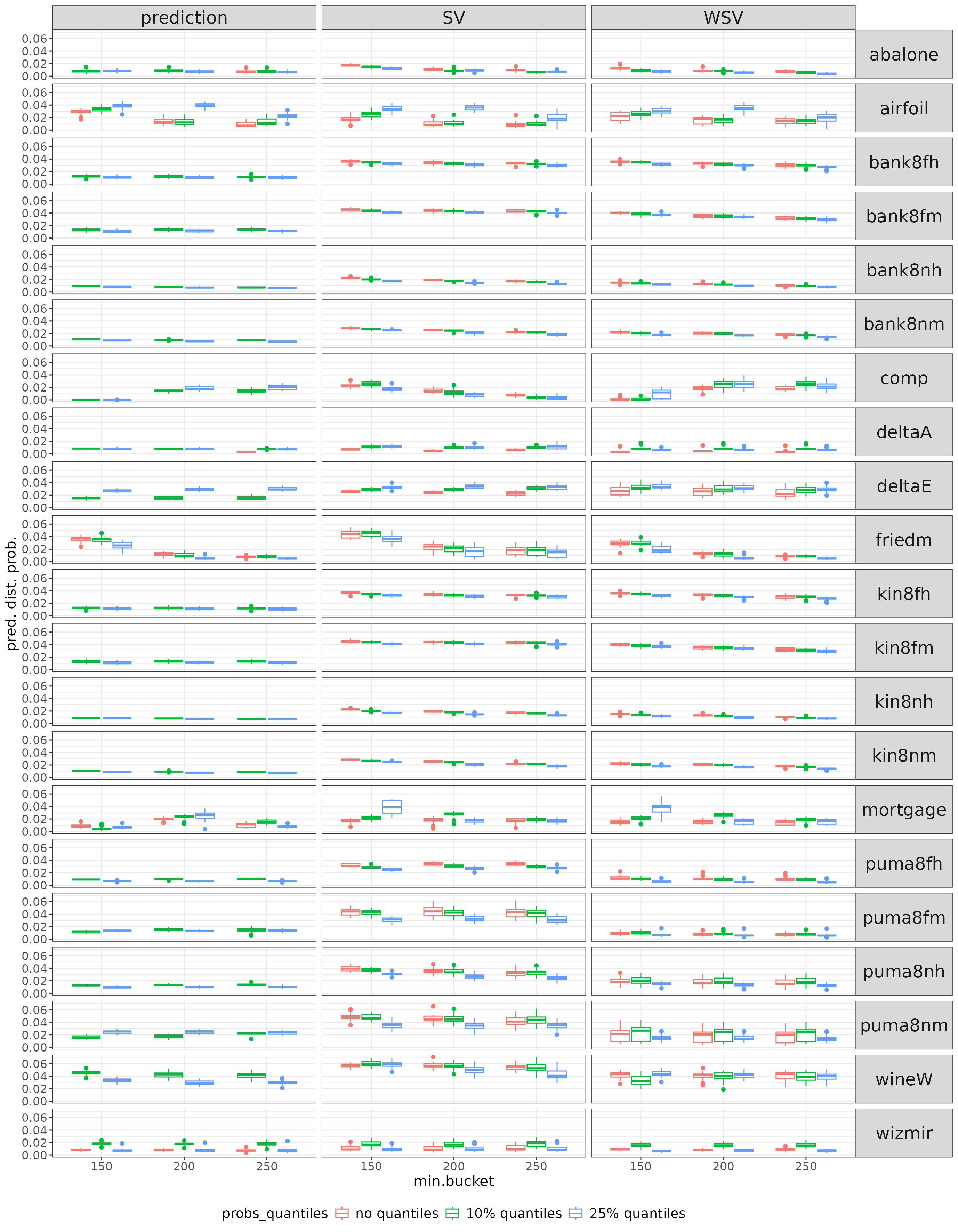}
	\caption{The prediction distance for the probability outcome is shown for the ART with CPS across 21 benchmark datasets. Each of the 20 repetitions was averaged over 10 folds. We varied the hyperparameters \emph{min.bucket} (x-axis), \emph{metric} (columns), and \emph{probs\_quantiles} (colors).}
	\label{appendix:fig:benchmark:hyperparameter_prediction_stability_probability}
\end{figure*}


\begin{figure*}[h!]
	\includegraphics[width=0.7\textwidth]{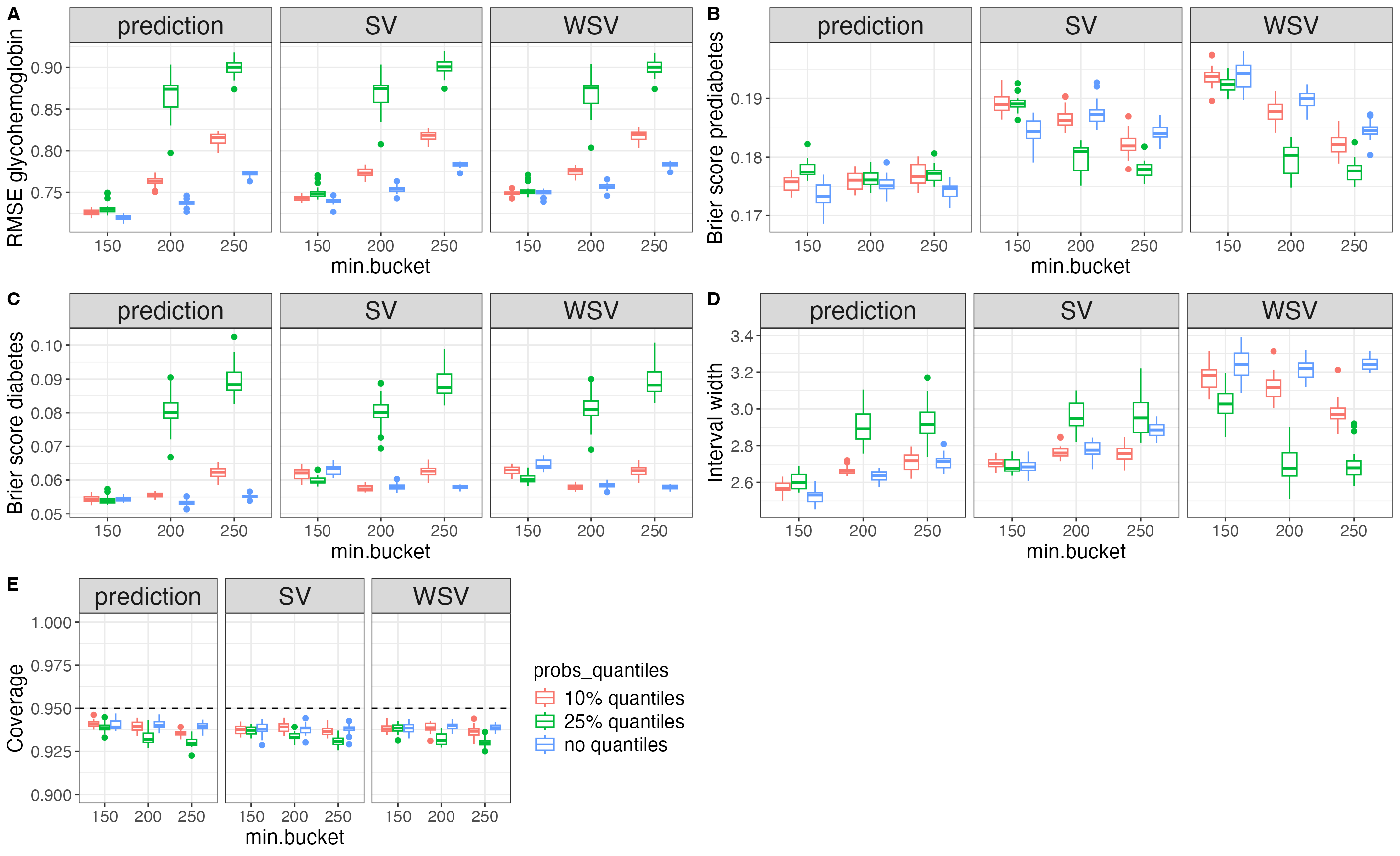}
	\caption{The prediction accuracy measures are shown for the ART with CPS across 20 repetitions. These include the RMSE (A) for predicting glycohemoglobin, the Brier score for predicting prediabetes (B) and diabetes (C), the width of
	the 95\% prediction interval (D) and the coverage (E) of the prediction interval. The results are averaged over the 10 folds. The variation of hyperparameter \emph{min.bucket} (x-axis), \emph{probs\_quantiles} (colors), and distance \emph{metric} (columns) are shown.
	\label{appendix:fig:accuracy}} 
\end{figure*}

\begin{figure*}[h!]
	\includegraphics[width=0.5\textwidth]{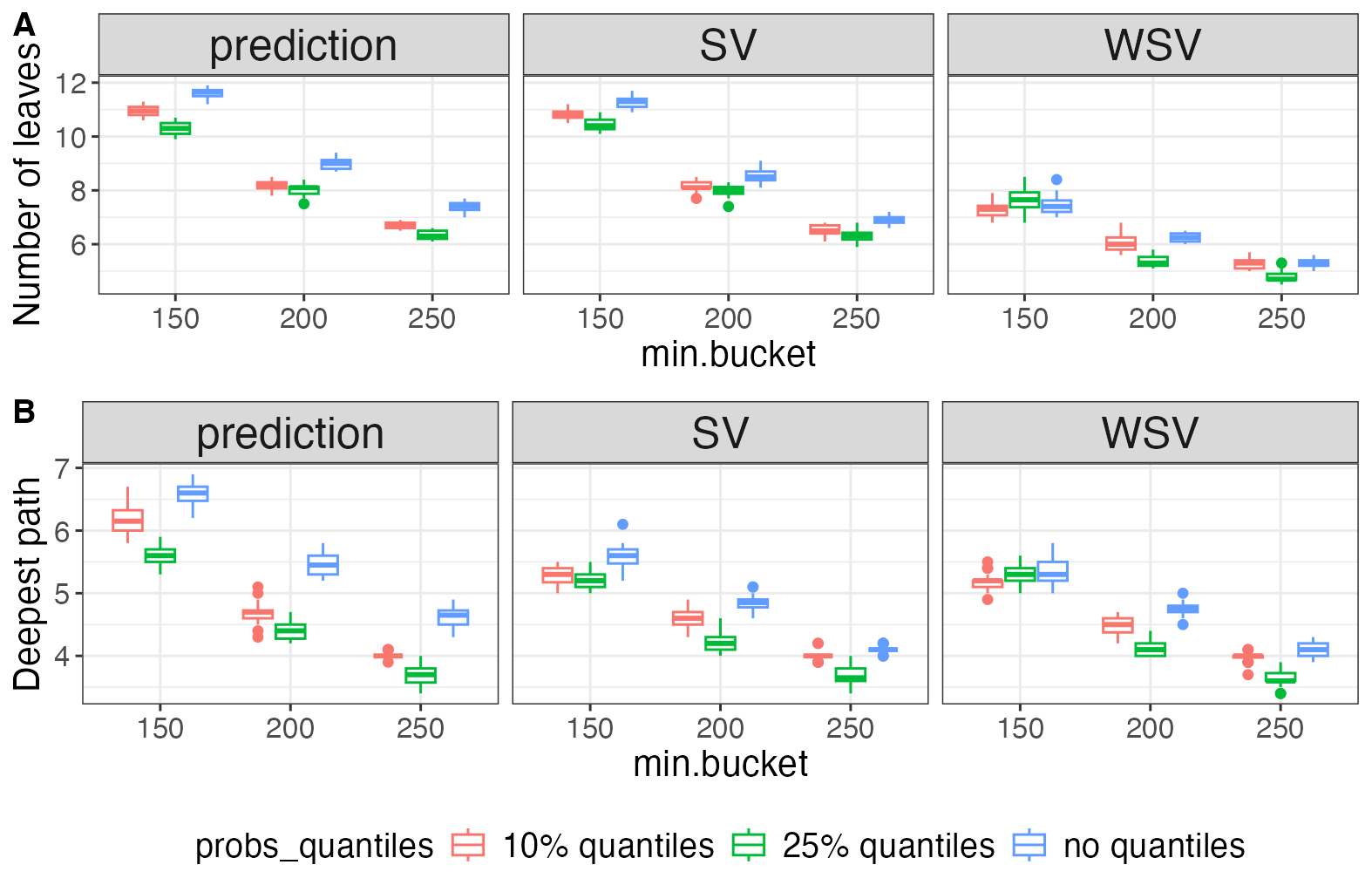}
	\caption{The number of leaves (A) and the deepest path (B) are shown for the ART with CPS across 20 repetitions. The results are averaged over the 10 folds. The variation of hyperparameter \emph{min.bucket} (x-axis), \emph{probs\_quantiles} (colors), and distance \emph{metric} (columns) are shown.
	\label{appendix:fig:interpretability_depth_leaves}} 
\end{figure*}

\begin{figure*}[h!]
	\includegraphics[width=0.7\textwidth]{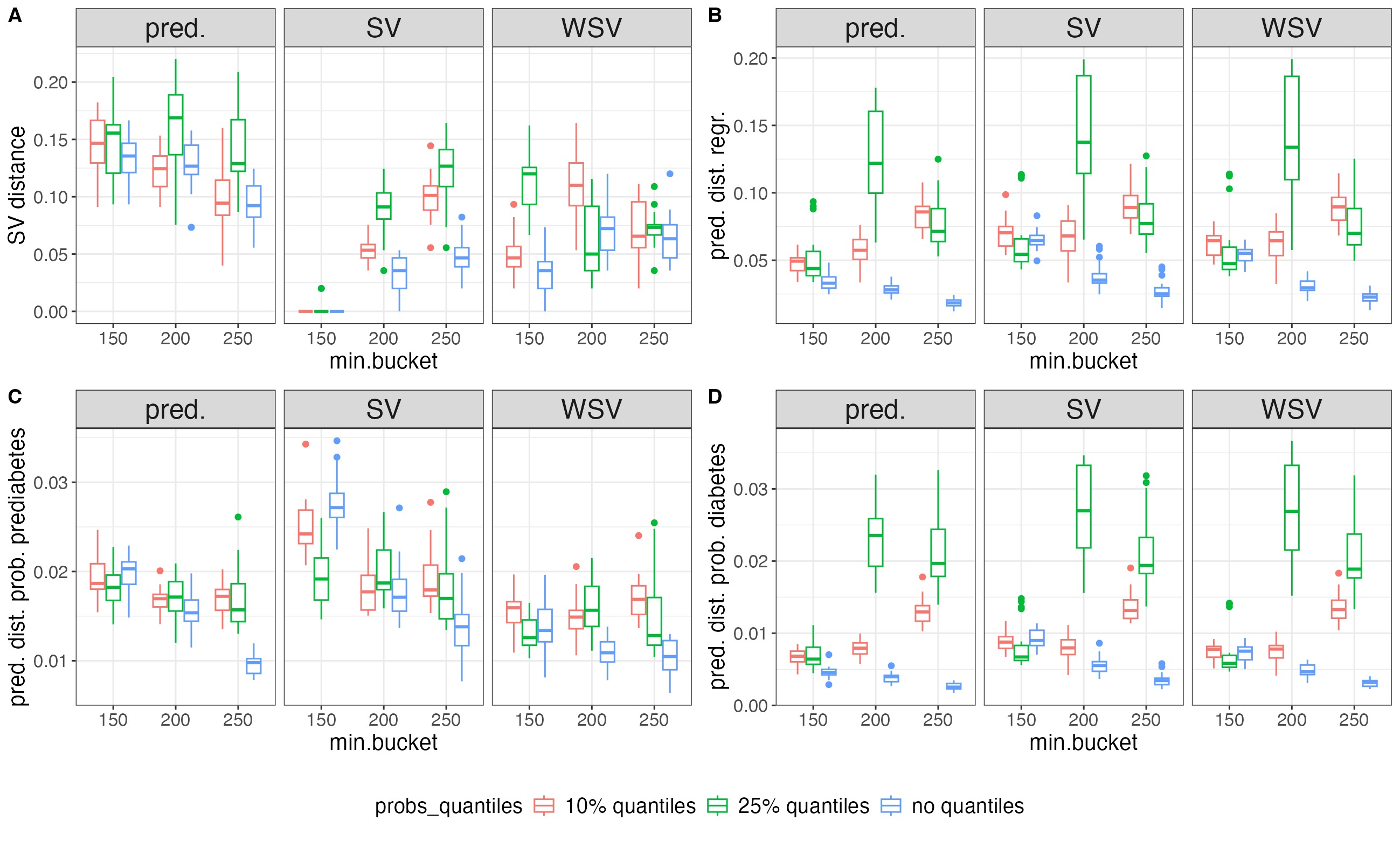}
	\caption{The splitting variable distance (A) and the prediction distance for the regression outcome (B), the probability outcome for prediabetes (C), and the probability outcome for diabetes (D)  are shown for the ART with CPS across 20 repetitions. The results are averaged over the 10 folds. The variation of hyperparameter \emph{min.bucket} (x-axis), \emph{probs\_quantiles} (colors), and distance \emph{metric} (columns) are shown.
	\label{appendix:fig:stability}} 
\end{figure*}

\begin{figure*}[h!]
	\includegraphics[width=0.7\textwidth]{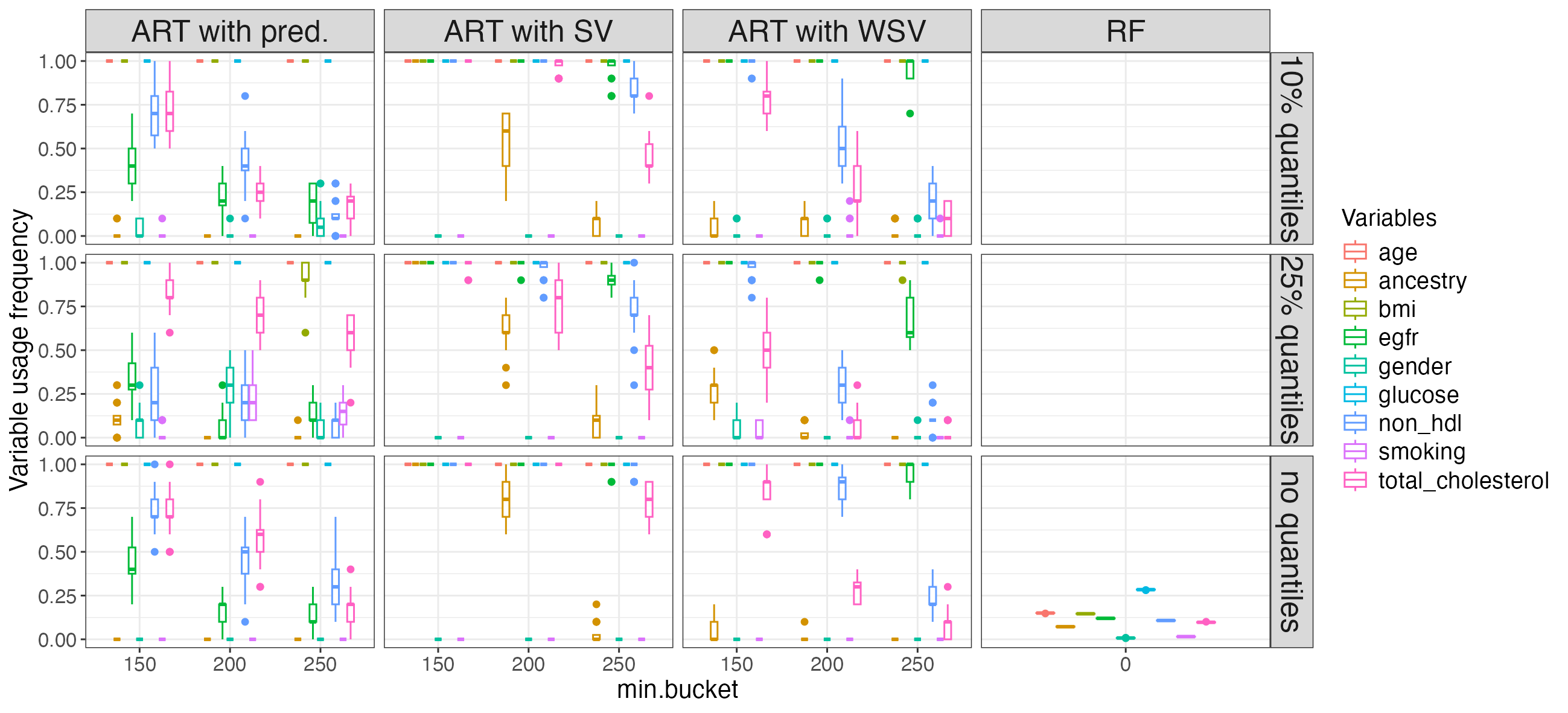}
	\caption{The frequency of each variable (colors) used as split variables is shown for the ART with CPS and RF across 20 repetitions. The results are averaged over the 10 folds. The variation of ART hyperparameter \emph{min.bucket} (x-axis), \emph{probs\_quantiles} (rows), and distance \emph{metric} (columns) are shown.
	\label{appendix:fig:stability_variable_usage}} 
\end{figure*} 

\end{document}